\documentclass[10pt,twocolumn,letterpaper]{article}

\usepackage[pagenumbers]{cvpr} % To force page numbers, e.g. for an arXiv version

\usepackage{url}
\usepackage{multirow}
\usepackage{enumitem}
\usepackage{stmaryrd}
\usepackage{array}
\usepackage{threeparttable}
\usepackage{blindtext}
\usepackage{amssymb}

\usepackage[ruled,vlined,linesnumbered,longend]{algorithm2e}
\usepackage{svg}
\usepackage{placeins}

\usepackage{soul}
\usepackage[utf8]{inputenc}
\usepackage{graphicx} % 导入包
\usepackage{amsmath}
\usepackage{booktabs}
\usepackage{lipsum}

\usepackage{threeparttable}
\usepackage{makecell}
\usepackage{booktabs}
\usepackage{multirow}
\usepackage{marvosym}
\usepackage{tikz}
\usepackage{amsfonts,amssymb}
\usepackage{longtable}
\usepackage{float}
\usepackage{placeins}
\usepackage[table]{xcolor}

\newcommand{\hide}[1]{}

\newcommand{\model}{KPM}
\newcommand{\bench}{KPM-Bench}
\newcommand{\moext}{MoPE}
\newcommand{\benchcap}{KPM-Cap}
\newcommand{\benchqa}{KPM-QA}
\newcommand{\benchhall}{KPM-HA}

\newcommand\blfootnote[1]{%
  \begingroup
  \renewcommand\thefootnote{}\footnote{#1}%
  \addtocounter{footnote}{-1}
  \endgroup
}
\usepackage{amsmath,amsfonts,bm}

\def\eqref#1{equation~\ref{#1}}
\def\1{\bm{1}}

\DeclareMathAlphabet{\mathsfit}{\encodingdefault}{\sfdefault}{m}{sl}
\SetMathAlphabet{\mathsfit}{bold}{\encodingdefault}{\sfdefault}{bx}{n}

\definecolor{cvprblue}{rgb}{0.21,0.49,0.74}
\usepackage[pagebackref,breaklinks,colorlinks,allcolors=cvprblue]{hyperref}

\title{KPM-Bench: A Kinematic Parsing Motion Benchmark for Fine-grained Motion-centric Video Understanding}

\author{
    \bf Boda Lin$^{1*}$,\qquad Yongjie Zhu$^{1\dagger}$,\qquad Xiaocheng Gong$^1$,\qquad Wenyu Qin$^{1}$\\
    {\bf Meng Wang}$^1$ \\
    $^1$Kuaishou Technology\\
  \texttt{\{linboda, zhuyongjie\}@kuaishou.com}
   }

\begin{document}

\twocolumn[{
\renewcommand\twocolumn[1][]{#1}
\maketitle
\begin{center}
    \centering
    \includegraphics[width=\textwidth]{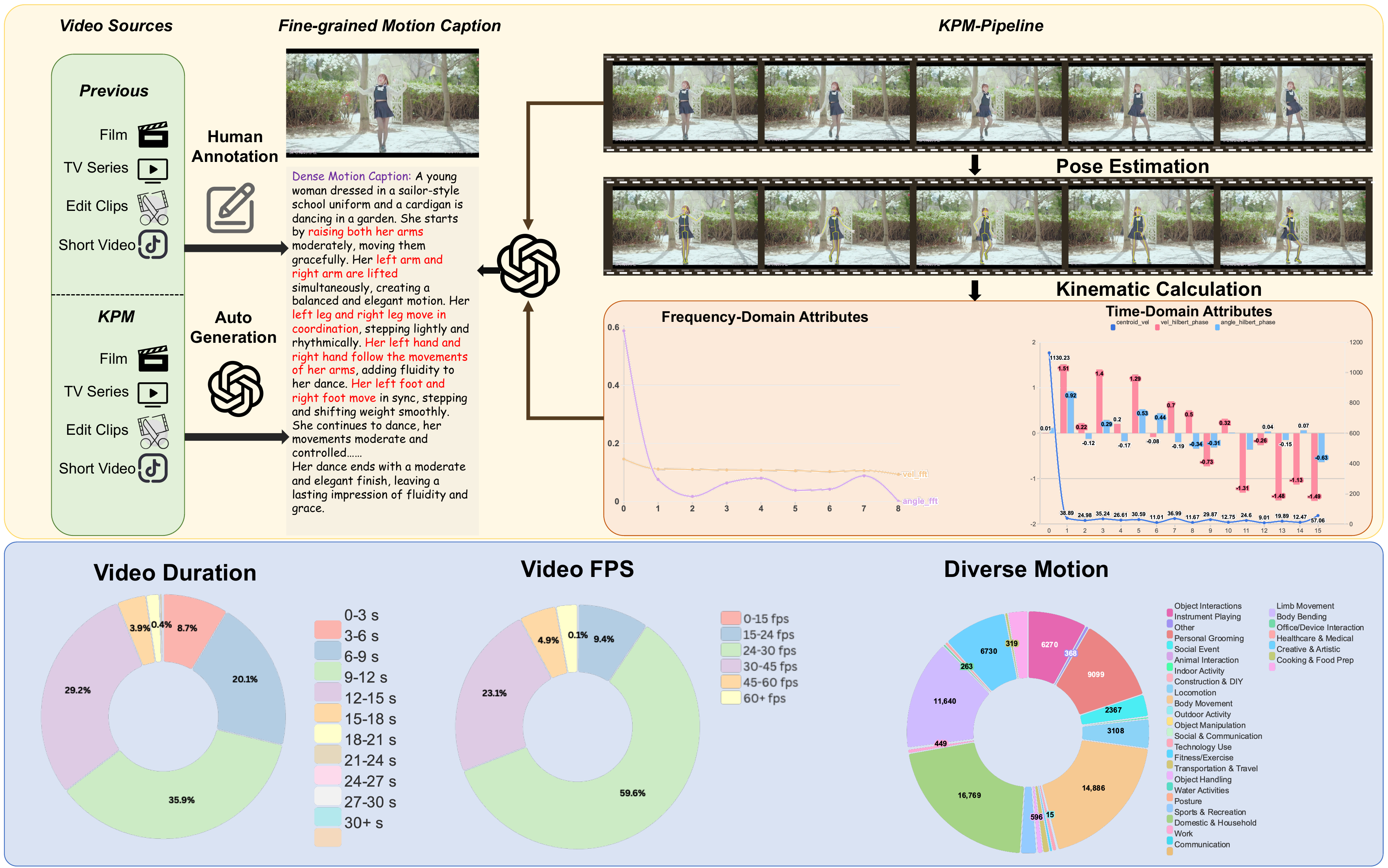}
     \captionof{figure}{Details of KPM-Bench. In contrast to previous approaches that rely solely on manual annotation of fine-grained motion-centric captions or directly generate annotations using GPT, our method enhances the automatic annotation process by first performing pose estimation on the video and then computing relevant physical attributes. This integration of motion analysis enables the generation of motion captions that can effectively unfold the detailed progression of complex actions.
}
    \label{fig:teaser}
\end{center}
}]

\blfootnote{$^*$This work was conducted during the author's internship at Kling Team, Kuaishou Technology. $^\dagger$Project learer.}

%%%%%%%%% ABSTRACT

\begin{abstract}
Despite recent advancements, video captioning models still face significant limitations in accurately describing fine-grained motion details and suffer from severe hallucination issues. 
These challenges become particularly prominent when generating captions for motion-centric videos, where precise depiction of intricate movements and limb dynamics is crucial yet often neglected.
To alleviate this gap, we introduce an automated annotation pipeline that integrates kinematic-based motion computation with linguistic parsing, enabling detailed decomposition and description of complex human motions. 
Based on this pipeline, we construct and release the Kinematic Parsing Motion Benchmark (KPM-Bench), a novel open-source dataset designed to facilitate fine-grained motion understanding. 
KPM-Bench consists of (i) fine-grained video-caption pairs that comprehensively illustrate limb-level dynamics in complex actions, (ii) diverse and challenging question-answer pairs focusing specifically on motion understanding, and (iii) a meticulously curated evaluation set specifically designed to assess hallucination phenomena associated with motion descriptions.
Furthermore, to address hallucination issues systematically, we propose the linguistically grounded Motion Parsing and Extraction (MoPE) algorithm, capable of accurately extracting motion-specific attributes directly from textual captions. 
Leveraging MoPE, we introduce a precise hallucination evaluation metric that functions independently of large-scale vision-language or language-only models. By integrating MoPE into the GRPO post-training framework, we effectively mitigate hallucination problems, significantly improving the reliability of motion-centric video captioning models.
\end{abstract}

\section{Introduction}
\label{sec:intro}

Video captioning is a fundamental task in computer vision and supports downstream applications such as video generation~\cite{wang2023videocomposer, villegas2023phenaki}. 
Recent advances in pre-training have brought Vision-Language Models (VLMs) that integrate vision encoders with large language models (LLMs)~\cite{Qwen-VL, Qwen2VL, yao2024minicpm, li2024videochat, lin-etal-2024-video, maaz-etal-2024-video}. 
These models now dominate the field and perform well on generic video descriptions. 
However, although these VLMs capture high-level content such as backgrounds, characters, and coarse motions, they still struggle to produce fine-grained motion descriptions. 
In practice, complex human movements are often reduced to broad summaries rather than presented as structured, limb-level analyses. 
Besides, motion hallucinations, namely non-existent or inaccurate motion details, remain common and undermine the reliability of current VLM-generated captions.

In this paper, we aim to tackle these two critical limitations: the lack of fine-grained motion descriptions and the high hallucination rate observed in motion-centric video captioning.

We first address the problem of insufficient fine-grained motion descriptions. 
A straightforward solution is supervised fine-tuning (SFT) on video-caption datasets containing detailed annotations. Unfortunately, current datasets rarely provide sufficient detail on limb-level motion decomposition at scale. Although datasets such as MotionBench~\cite{hong2024motionbench} deliver high-quality fine-grained annotations, they depend entirely on costly manual labeling, substantially restricting dataset scale.

To overcome this limitation, we propose an automated annotation pipeline integrating kinematic-based motion decomposition and structured linguistic parsing. Specifically, based on Screw Theory~\citep{murray2017mathematical} and Chasles' Theorem~\citep{rodrigues1840lois}, we systematically decompose human motion into two orthogonal components: \textbf{position translation} and \textbf{postural transformation}. 
In the temporal domain, we quantify these motions by computing translational velocities and angular velocities of skeletal joints. 
Furthermore, in the frequency domain, we employ Fast Fourier Transform (FFT) analysis to measure rhythmic variations in motion intensity, effectively distinguishing between vigorous and subtle movements. 
Inspired by linguistic frameworks such as case grammar~\cite{fillmore1968case} and motion event theory~\cite{talmy1985lexicalization}, we also define a structured Parsing Motion Representation (\textbf{PaMoR}), enabling systematic and comprehensive coverage of motion-related semantic components through template-based linguistic realization.

Utilizing this pipeline, we build a new large-scale benchmark dataset, \textbf{KPM-Bench}, consisting of over $75$k video-caption pairs with detailed motion descriptions. 
The benchmark additionally includes $38$k carefully designed complex question-answer (QA) pairs tailored explicitly for motion comprehension assessment and a rigorously annotated hallucination evaluation set specifically constructed to challenge and measure motion-related hallucinations.

Next, we address the hallucination issue prevalent in VLM-generated motion captions. 
We begin by systematically classifying hallucination types unique to motion descriptions, then introduce a novel linguistically grounded algorithm—Motion Parsing and Extraction (\textbf{\moext})—to precisely identify motion attributes described in captions. 
Based on the proposed \moext, we design a dedicated reward function integrated into the Group Relative Policy Optimization (GRPO)~\cite{shao2024deepseekmath} training process. 
Through this targeted post-training strategy, we substantially mitigate hallucination occurrences, significantly enhancing caption reliability.
Finally, we establish a new precise hallucination measurement metric independent of external LLMs and VLMs.

Collectively, these contributions offer a comprehensive methodological and benchmarking approach that substantially improves fine-grained motion description accuracy and reduces hallucination phenomena in motion-centric video captioning, thus facilitating more precise and reliable video understanding.
\section{Related Work}
\label{sec:related}

\subsection{Fine-grained Video Captioning with VLMs}
Recent advances in video-language models have explored granular motion description through various paradigms. 
MotionBERT~\cite{motionbert} pioneers joint training of motion encoders and language decoders using human pose sequences, but lacks video context integration. 
MotionLLM~\cite{chen2024motionllm} also use the VQ-VAE to encode the motion and propose MLP-based V-L translator to project the motion representations and video representations into the language space.
MotionBench~\cite{hong2024motionbench} provides 5K high-quality, fine-grained motion-related video captions, all manually annotated. This ensures a strong level of detail and accuracy, making it a valuable resource for evaluating motion understanding in video captioning models.

\subsection{Hallucination Evaluation}
Evaluating hallucination in video captioning, particularly for fine-grained motion-centric descriptions, presents a significant challenge. Current approaches to measure hallucination in video captions, primarily fall into three categories. Manual evaluation~\cite{cui2023holisticanalysishallucinationgpt4vision} by humans provides detailed understandings but is costly and susceptible to annotator biases and inconsistencies.  Another approach involves matching generated captions against predefined, granular benchmarks. While benchmarks like CHAIR~\cite{rohrbach2018object}, which utilizes the MSCOCO dataset~\cite{lin2014microsoft} to provide detailed object annotations for image captioning, similar comprehensive and widely accepted fine-grained standards for video captioning, especially concerning the intricate details of motion, are scarce. Lastly, leveraging Vision Language Models (VLMs) or Large Language Models (LLMs) as automated evaluators~\cite{guan2024hallusionbench, liu2023mitigating, wang2023evaluation, zhai2023halle} is an emerging trend. However, these evaluator models are themselves prone to hallucination, potentially introducing their own biases, and their performance can be highly sensitive to prompt phrasing and structure. 

\section{KPM-Bench: Kinematic Parsing Motion Benchmark}
\label{sec:bench}
\subsection{Kinematic Motion Computation}
\label{sec:phy}

\begin{figure*}[t]
\centering
\includegraphics[width=1.0\linewidth]{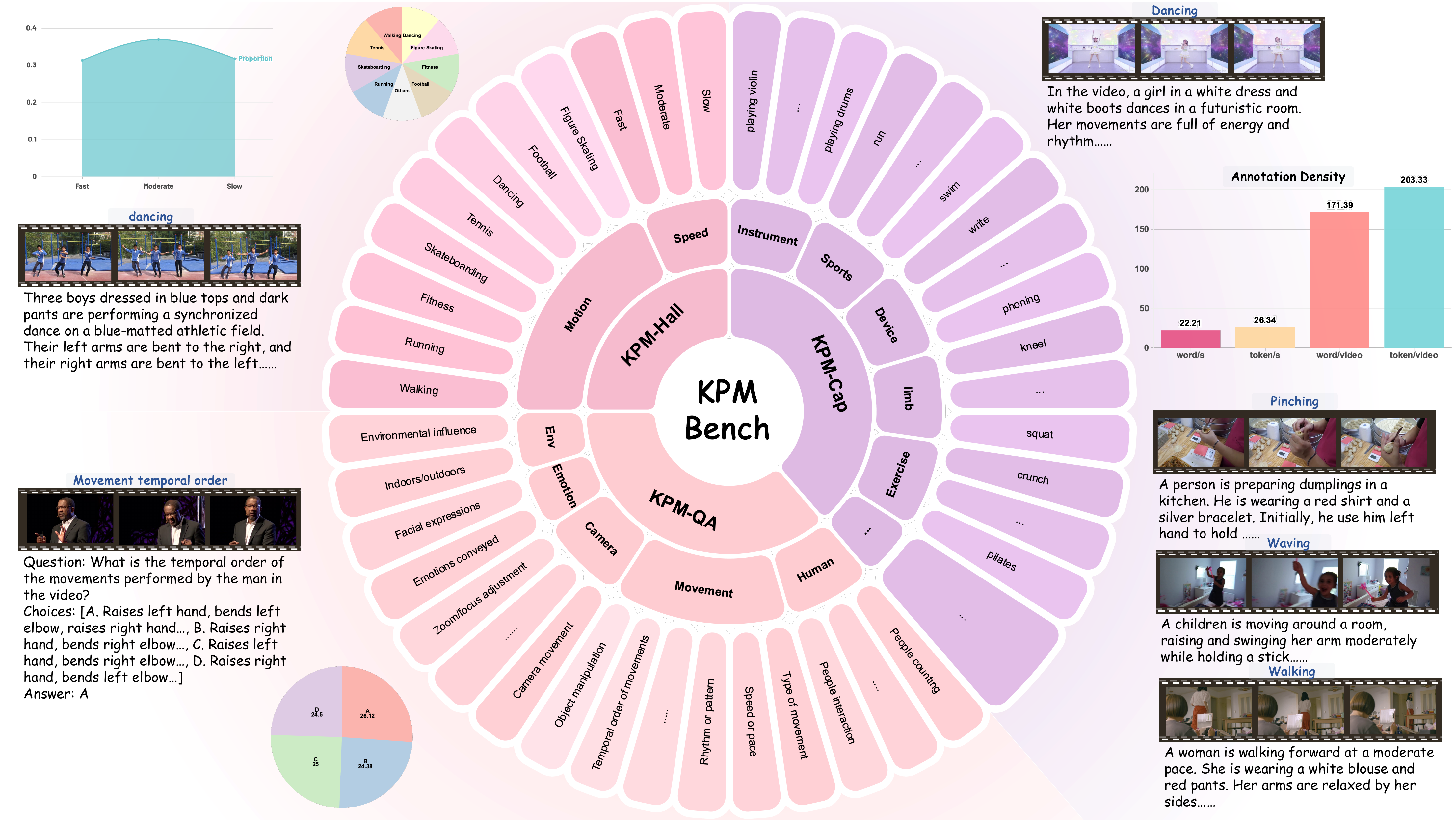}
\caption{
The statistics details of \bench.
The \bench~ contains three subsets: \benchcap, \benchqa, and \benchhall.
In addition to visualizing the cases and the category distribution of each subset, we also show the annotation density of the \benchcap, the option distribution statistics of the \benchqa, and the motion type statistics of the \benchhall.
}
\label{fig:bench:sunburst}
\end{figure*}
Motion inherently consists of two physical dimensions: position translation (spatial displacement) and postural transformation (limb movement), measurable via velocity and angular velocity respectively. 
To further capture rhythmic nuances, we integrate frequency-domain analysis.

\textbf{Time-domain Analysis.} We calculate the velocity of each skeletal point using the following formula:
\begin{equation}
v_k(t)=\frac{1}{\Delta t}\big\lVert
(\hat X_k^t,\hat Y_k^t,\hat Z_k^t)-(\hat X_k^{t-1},\hat Y_k^{t-1},\hat Z_k^{t-1})
\big\rVert_2
\end{equation}

Where $\hat{X}_k^t$, $\hat{Y}_k^t$, and $\hat{Z}_k^t$ denote the horizontal, vertical, and depth coordinates of the $k$-th skeletal point at frame $t$, respectively, and $\Delta t$ denotes the time interval between consecutive frames. 
After computing the velocities $v_k(t)$ for all skeletal points, we obtain the center-of-mass velocity $v_{\mathrm{cm}}(t)$ as

\begin{equation}
    v_{cm}(t) = \frac{1}{K}\sum_{k=1}^{K} v_k(t)
\end{equation}
Then we use the cosine formula to calculate the projection angles of joints.
\begin{equation}
    \eta_j^t = \arccos\left(\frac{\langle LB_j^t, RB_j^t \rangle}{\|LB_j^t\|\|RB_j^t\|}\right)
\end{equation}
Where, $LB_j^t$ and $RB_j^t$ represent two limb segments forming the angle at the 
$j$-th joint.
We adopted the COCO-Wholebody 133-skeletal points representation~\cite{jin2020whole,xu2022zoomnas}, under which a total of 10 joint angles can be calculated and the details regarding the joint angles are shown in Section B of Appendix.

After calculating all joint angles $\eta$, the rotational angular velocity $\omega$ for each joint angle and the average angular velocity can be computed using the following formula:
\begin{equation}
    \omega_j(t) = \frac{\eta_j^t - \eta_j^{t-1}}{\Delta t}, \quad \bar{\omega} = \frac{1}{J}\sum_{j=1}^{J}\bar{\omega}_j
\end{equation}

\textbf{Frequency-domain Analysis.} First, we apply a Faster Fourier Transform~(FFT)~\cite{dft1979} to the velocity signals $v[t]$ and angular velocity signals $\omega[t]$ in the time domain.
\begin{equation}
    V[k] = \sum_{t=0}^{T-1} v[t] e^{-j\frac{2\pi}{T}kt}, \quad 
    \Omega[k] = \sum_{t=0}^{T-1} \omega[t] e^{-j\frac{2\pi}{T}kt}
\end{equation}
Based on the motion spectrum, we can further calculate the following kinematic quantities: the total energy $E$ of the motion signal, the proportion of high-frequency components $P$, and the standard deviation of the spectrum $\sigma$.
The formulas for calculating these kinematic quantities are as follows:
\begin{gather}
    E_V = \sum_{k=0}^{T-1} V[k]^2,\quad E_\Omega = \sum_{k=0}^{T-1}\Omega[k]^2\\[6pt]
    P_V = \frac{\sum_{k=\delta +1}^{T-1}V[k]^2}{\sum_{k=0}^{T-1}{V[k]^2}}, \quad
    P_\Omega = \frac{\sum_{k=\delta +1}^{T-1}\Omega[k]^2}{\sum_{k=0}^{T-1}{\Omega[k]^2}}\\[6pt]
    \sigma_V = \sqrt{\frac{1}{T}\sum_{k=0}^{T-1}(V[k]-\bar{V})^2}, \quad
    \sigma_\Omega = \sqrt{\frac{1}{T}\sum_{k=0}^{T-1}(\Omega[k]-\bar{\Omega})^2}
\end{gather}

\subsection{Linguistic Parsing Representation}
Inspired by Case Grammar~\cite{fillmore1968case} and the Motion Event Framework~\cite{fillmore2006frame}, we propose a linguistic schema named \textbf{Pa}rsing-based \textbf{Mo}tion Event \textbf{R}epresentation (\textbf{PaMoR}), specifically crafted for detailed motion-centric video captioning. PaMoR systematically captures complex human motion through a hierarchical classification coupled with standardized attribute representation, transcending simple attribute enumeration by offering a structured, parsing-oriented semantic framework.
The case of PaMoR are shown in Section B of Appendix.

\textbf{Hierarchical Motion Classification.} PaMoR classifies motion into three distinct yet interconnected semantic levels: \textit{individual-level}, \textit{limb-level}, and \textit{distal-level}. The \textbf{individual-level} delineates global bodily movements, such as walking or jumping, reflecting overall spatial trajectories. The \textbf{limb-level} encapsulates coordinated movements of major body segments, including arm swings or torso bends. The \textbf{distal-level} highlights subtle articulations of extremities like finger gestures or head tilts, often crucial for expressive nuance.

\textbf{Standardized Motion Representation.} To precisely encode motion events, PaMoR defines eight core attributes:
(1) \textbf{MI (Motion Predicate)} identifies the core kinetic action;
(2) \textbf{AE (Agentive Entity)} denotes the motion initiator;
(3) \textbf{PE (Patientive Entity)} captures entities impacted by the motion;
(4) \textbf{MM (Magnitude Modifier)} quantifies motion intensity;
(5) \textbf{DI (Direction Indicator)} specifies directional orientation;
(6) \textbf{AQ (Agent Qualifiers)} provides descriptive properties of the agent;
(7) \textbf{PQ (Patient Qualifiers)} describes detailed characteristics of the patient.

By integrating these structured attributes, PaMoR ensures nuanced semantic coherence and interpretability. 
The structured parsing nature of this representation not only clarifies complex motion sequences but also facilitates consistent, detailed linguistic alignment with visual content, thus validating its standalone naming and application as a sophisticated motion event representation framework.

\begin{figure*}[t]
\centering
\includegraphics[width=1.0\linewidth]{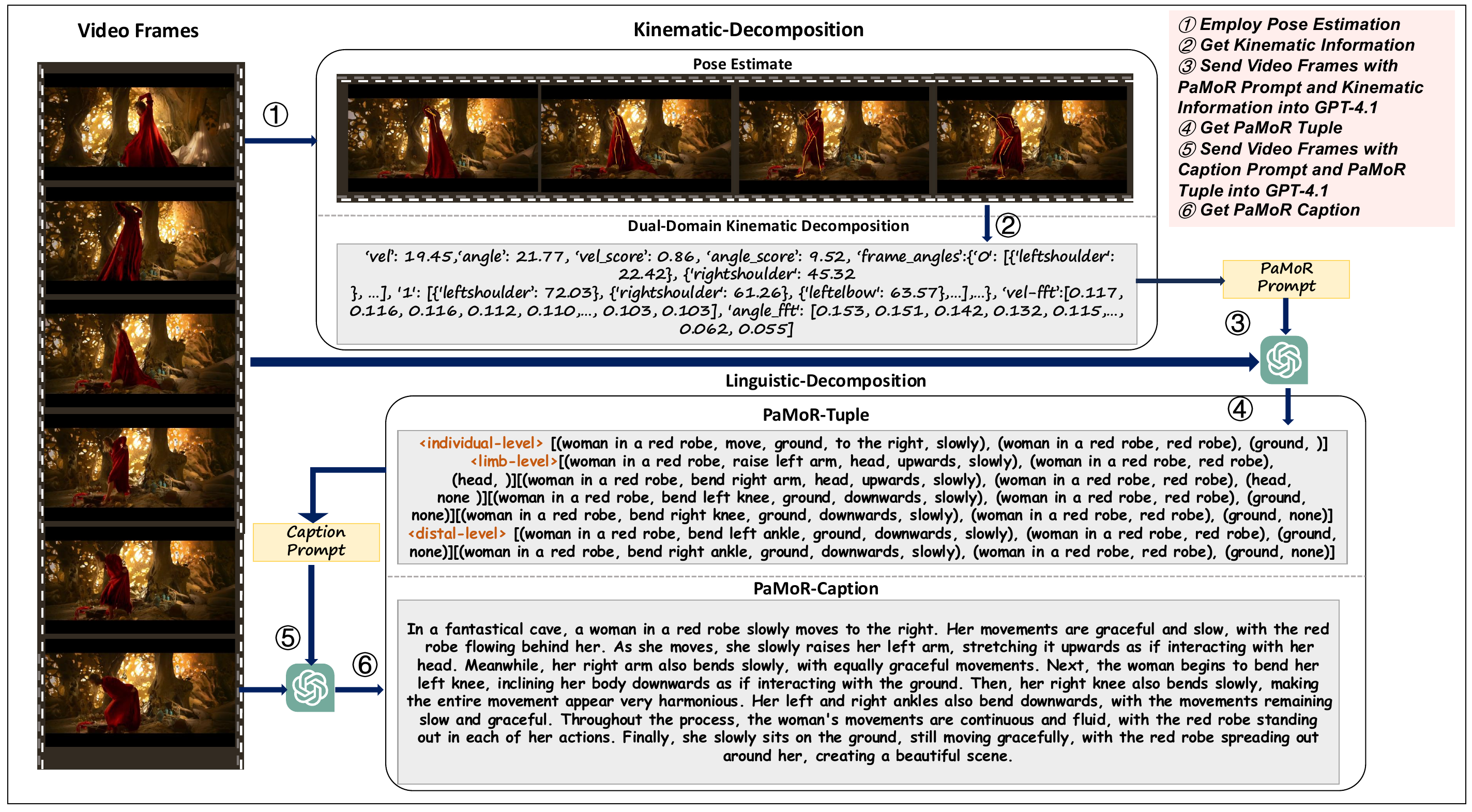}
\caption{
The pipeline of dataset construction.
First, the video frames are processed through 3D pose estimation and kinematic calculation to obtain kinematic-related numerical attributes. 
Then, these kinematic attributes and the video frames are fed into GPT to obtain PaMoR-Tuple format annotations, which are then further expanded into dense captions.
}
\label{fig:data:pipeline}
\end{figure*}

\subsection{Automatic Annotation Pipeline}
\label{sec:sub:data:pipeline}

In this section, we provide a detailed introduction to the proposed \model~dataset and the corresponding standardized construction process.

As illustrated in Figure~\ref{fig:data:pipeline}, the process begins by collecting videos from various sources and segmenting them into clips.
As shown in Figure~\ref{fig:teaser}, we keep the length of most video clips under 30 seconds. 
This is because we want to generate motion descriptions with a sufficiently high annotation density, while excessively long videos will lose details during motion due to limitations in sampling fps and frame number.
The preprocessed video frames undergo pose estimation via RTMPose3D~\cite{jiang2023rtmpose}, a real-time model for 3D wholebody pose estimation.

Following the pose annotations, we implement the kinematic analysis framework detailed in Section~\ref{sec:phy}.
As shown in Figure~\ref{fig:data:pipeline}, this allows us to extract detailed body movement information for person in the video, which includes velocity, angular velocity, velocity score, angular velocity score, velocity FFT sequence, angular velocity FFT sequence, and the angle of each joint for every frame. 
Such fine-grained motion analysis generates a comprehensive numerical dictionary for each frame, capturing the intricate dynamics of movement.
Subsequently, the kinematical attributes are embedded into prompts using the template shown in Section~B of the Appendix. 
Six-stage prompting with GPT-4.1 generates PaMoR-format annotations.
Finally, the PaMoR annotations and original videos undergo multi-modal fusion through the pipeline in Figure~\ref{fig:data:pipeline}, where GPT-4.1 produces dense captions.
Different from directly employ GPT-4.1 to generate video captions using single prompt, our pipeline facilitate the generation of dense video captions that provide a detailed representation of the motion process.

\subsection{Data Statistics}
The \bench~ is composed of three components: \benchcap~which includes $75$k videos with dense fine-grained motion-centric video caption, \benchqa~which includes $38$k videos with question-answer pairs, \benchhall~which includes $215$ videos selected from $20$ movement types.

\benchcap~consists of fine-grained motion captions generated from a diverse set of video sources. 
As illustrated in Figure~\ref{fig:bench:sunburst}, \benchcap~achieves an annotation density of $22.21$ words per second and $171.39$ words per video, providing highly detailed descriptions of fine-grained movements within the videos. 
Moreover, \benchcap~demonstrates strong capacity for unpacking complex actions—such as dancing, playing sports, or surfing—by elaborating the motion process down to the level of how individual body parts move and interact.

\benchqa~is composed of complex, motion-related questions. As illustrated in Figure~\ref{fig:bench:sunburst}, we design $5$ categories encompassing $38$ types of complex question templates, which decompose motion from multiple perspectives, including motion attributes, interaction states, body poses, and contextual background. Corresponding answers to these questions are generated using GPT‑4.1 in combination with the video content and caption annotations.
The details of QA templates and QA construction are shown in Section C of Appendix.

\benchhall~consists of video captions that are prone to hallucination. We conducted a detailed statistical analysis and classification of motion captions and selected 215 videos covering a wide range of motion categories. For these videos, we carefully designed fine-grained motion captions through manual annotation.

\subsection{Caption Comparison}
\begin{figure*}[t]
\centering
\includegraphics[width=1.0\linewidth]{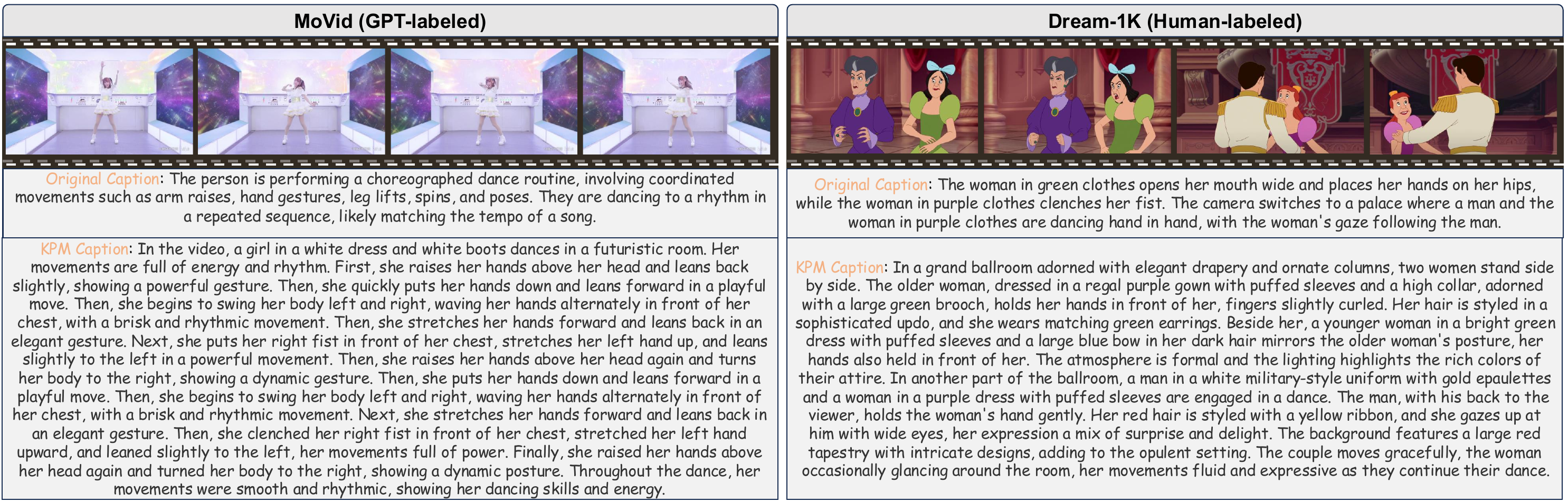}
\caption{
Comparison cases of caption annotation performance using the our method on MoVid~\cite{chen2024motionllm} and Dream-1K~\cite{yuan2025tarsier2}.
}
\label{fig:data:bench:case}
\end{figure*}

As shown in the Figure~\ref{fig:data:bench:case}, we compared two Video-Caption datasets, one annotated with GPT and the other with human annotations. 
We re-annotated the videos using the baseline from the section~\ref{sec:sub:data:pipeline}.
Our re-annotated captions have a higher annotation density and provide richer descriptions of motion processes and detailed actions.

\section{Post-training}
\label{sec:post}
\subsection{Hallucinations Classification for Motion Caption}
We conducted extensive analyses of motion-related hallucinations in video captioning, and the results indicate that the most frequently observed hallucinations can be broadly categorized into three types: (1) adding descriptions of actions that do not exist in the video; (2) scrambling the temporal order of actions; and (3) incorrectly describing the direction or orientation of actions.

\subsection{\moext: Motion Parsing and Extraction Algorithm}
We propose a Motion Parsing and Extraction (\moext) algorithm to distill fine-grained actions and their temporal order from textual video captions. Our method uniquely combines Abstract Meaning Representation (AMR)~\cite{banarescu2013abstract} for deep semantic parsing and dependency parsing (DP)~\cite{mel1988dependency} for syntactic precision.
The pseudo code of the entire extraction algorithm is shown in the Section B of Appendix.

\textbf{Dual Parsing.} Captions are first preprocessed by coreference resolution, then parsed via AMR (semantic graphs) and DP (syntactic relations).

\textbf{Action Extraction.} Dynamic actions are identified from AMR predicates marked with PropBank's~\cite{kingsbury-palmer-2002-treebank} \texttt{-01} suffix, excluding static verbs. Semantic roles are then extracted for each action.

\textbf{Syntactic Alignment.} AMR actions are aligned with DP verb tokens. Subjects, objects, and modifiers are refined using syntactic dependencies (\texttt{nsubj}, \texttt{dobj}, \texttt{advmod}).

\textbf{Attribute Fusion.} Semantic roles (from AMR) and syntactic details (from DP) are fused for comprehensive attribute coverage.
\begin{table*}[htbp]
\centering
\scalebox{0.66}{
\begin{tabular}{ccc|cccccc|cc}
\toprule
\textbf{Model} &  \textbf{Utilized-LLM} & \textbf{Size} & \textbf{B-1$\uparrow$} & \textbf{B-4$\uparrow$} & \textbf{R-L$\uparrow$} & \textbf{B-Score$\uparrow$} & \textbf{GPT-Score$\uparrow$} & \textbf{GPT-Acc$\uparrow$} & \textbf{GPT-Hall$\downarrow$} & \textbf{Mo-Hall$\uparrow$}\\
\hline
\rowcolor{gray!10}
\multicolumn{3}{l|}{\textit{Closed-source VLM-APIs}} & & & & & & & &\\
GPT-4.1~\cite{gpt-4-1} & - & - & $38.78$ & $9.99$ & $29.68$ & $93.73$ & $3.23$ & $62.41\%$& $2.32$ & $0.603$\\
Gemini-2.5Pro~\cite{genimi-2p5} & - & - & $32.22$ & $4.52$ & $25.52$ & $92.80$ & $3.11$ & $58.79\%$& $2.37$ & $0.595$\\
\hline
\rowcolor{gray!10}
\multicolumn{3}{l|}{\textit{Open-sourced Video-VLMs}} & & & & & & & &\\
InternVideo-2.5~\cite{wang2025internvideo2} & InternLM2.5~\cite{wu2024internlm2-5} & 8B &  $20.46$ & $3.05$ & $19.45$ & $86.56$ & $2.35$ & $30.70\%$& $2.48$ & $0.573$\\
MiniCPM-V2.6~\cite{yao2024minicpm} & Qwen2~\cite{yang2024qwen2technicalreport} & 8B & $24.39$ & $2.24$ & $20.06$ & $92.77$ & $2.52$ & $34.36\%$& $2.55$ & $0.541$\\
VideoLLaMA3~\cite{zhang2025videollama} & Qwen2.5 & 7B & $23.49$ & $3.18$ & $21.86$ & $85.88$ & $2.30$ & $28.18\%$ & $2.45$ & $0.562$\\
LLaVA-OneVision~\cite{li2024llavaov} & Qwen2 & 7B & $29.77$ & $3.24$ & $21.07$ & $93.74$ & $2.53$ & $31.55\%$& $2.71$ & $0.579$ \\
Tarsier2-Recap~\cite{yuan2025tarsier2}  & Qwen2 & 7B & $33.58$ & $5.61$ & $23.99$ & $93.62$ & $3.01$ & $43.51\%$& $\textbf{1.87}$ & $\textbf{0.621}$\\
Qwen2-VL~\cite{Qwen-VL} & Qwen2~\cite{yang2024qwen2technicalreport} & 7B & $21.60$ & $1.74$ & $19.19$ & $89.65$ & $2.60$ & $25.64\%$& $2.44$ & $0.563$\\
Qwen2.5-VL~\cite{Qwen2.5-VL} & Qwen2.5~\cite{yang2024qwen2-5} & 7B & $27.51$ & $2.83$ & $20.14$ & $91.23$ & $3.19$ & $54.63\%$ & $2.41$ & $0.585$ \\
\hline
\model~(w/o MoPE) & Qwen2.5 & 7B & $\textbf{43.64}$ & $\textbf{15.14}$ & $\textbf{34.11}$ &  $\textbf{96.57}$ & $3.39$ & $67.83\%$ & $2.03$ & $0.607$\\
\model & Qwen2.5 & 7B & $43.25$ & $14.77$ & $33.40$ & $96.51$ & $\textbf{3.46}$ & $\textbf{71.80}\%$ & $1.92$ & $0.619$\\

\bottomrule
\end{tabular}
}
\caption{Comparison of \model~with other closed-source and open-sourced VLMs on \benchcap~ and \benchhall. 
The B-1 means BLEU-1, B-4 means BLEU-4, R-L means Rouge-L, and B-Score means Bert-Score.
GPT-hall and Mo-Hall are metrics for detecting the number of sentence hallucinations and action hallucinations in the captions of the \benchhall, respectively.
}
\label{tab:exp:result:main}
\end{table*}

\textbf{Temporal Ordering.} Action sequences are inferred through explicit temporal markers, verb dependencies, AMR time roles, and implicit textual ordering. These cues are integrated into a directed action graph, topologically sorted to finalize action order.

\subsection{GRPO with \moext}

We integrate MoPE into the GRPO framework by designing a composite motion-aware reward that evaluates both the accuracy and coherence of generated captions. Specifically, we define the reward as a weighted combination of three components:

\begin{equation}
R(C_g, C_r) = w_a \cdot R_{\mathrm{action}} + w_o \cdot R_{\mathrm{order}} + w_d \cdot R_{\mathrm{direction}}
\end{equation}

Where $C_g$ and $C_r$ are the generated and reference captions, and $w_a$, $w_o$, $w_d$ are the weights satisfying $w_a + w_o + w_d =1$.

\paragraph{Action Accuracy}
We compute the F1-score over action concepts:
\begin{equation}
R_{\mathrm{action}} = \mathrm{F1}(\mathrm{Concepts}(A_g), \mathrm{Concepts}(A_r))
\end{equation}
Where $A_g = \mathrm{MoPE}(C_g)$ and $A_r = \mathrm{MoPE}(C_r)$.

\paragraph{Order Accuracy}
We calculate the proportion of correctly predicted pairwise relative orders among common actions:
\begin{equation}
R_{\mathrm{order}} = \frac{\#\{\text{correct relative pairs}\}}{\#\{\text{total gold pairs}\}}
\end{equation}

\paragraph{Direction Accuracy}
We measure the alignment of directional attributes among matched actions:
\begin{equation}
R_{\mathrm{direction}} = \frac{\#\{\text{correct directions}\}}{\#\{\text{relevant gold directions}\}}
\end{equation}

Where $\#$ means the number values.
This composite reward encourages models to generate captions with accurate actions, correct temporal ordering, and coherent motion attributes.

\section{Experiments}
\label{sec:exp}

\begin{table*}[htbp]
\centering
\scalebox{0.7}{
\begin{tabular}{ccc|cccccc}
\toprule
\textbf{Model} &  \textbf{Utilized-LLM} & \textbf{Size} &  \textbf{All$\uparrow$} & \textbf{Move$\uparrow$} & \textbf{Human$\uparrow$} & \textbf{Camera $\uparrow$} & \textbf{Emotion$\uparrow$} & \textbf{Environment$\uparrow$}  \\
\hline
\rowcolor{gray!10}
\multicolumn{3}{l|}{\textit{Closed-source VLM-APIs}} & & & & & & \\
% \rowcolor{blue!10}
GPT-4.1~\cite{gpt-4-1} & - & - & $84.41$ & $75.34$ & $77.62$ & $72.88$ & $88.13$ & $95.79$\\
% \rowcolor{blue!10}
Gemini-2.5Pro~\cite{genimi-2p5} & - & - & $81.55$ & $77.31$ & $76.22$ & $71.08$ & $87.93$ & $95.19$\\
\hline
% \rowcolor{green!10}
\rowcolor{gray!10}
\multicolumn{3}{l|}{\textit{Open-sourced Video-VLMs}} & & & & & &  \\
% \rowcolor{green!10}
InternVideo-2.5~\cite{wang2025internvideo2} & InternLM2.5~\cite{wu2024internlm2-5} & 8B &  $85.95$ & $83.70$ & $75.74$ & $77.42$ & $92.21$ & $94.22$\\
% \rowcolor{green!10}
VideoLLaMA3~\cite{zhang2025videollama} & QWen2.5 & 7B & $58.69$ & $57.25$ & $47.85$ & $54.39$ & $83.85$ & $61.26$\\
% \rowcolor{green!10}
LLaVA-OneVision~\cite{li2024llavaov} & QWen2.5 & 7B & $79.09$ & $73.41$ & $75.75$ & $68.42$ & $86.84$ & $93.91$\\
% \rowcolor{green!10}
Qwen2-VL~\cite{Qwen-VL} & Qwen2~\cite{yang2024qwen2technicalreport} & 7B & $76.39$ & $75.92$ & $58.19$ & $61.40$ & $85.58$ & $81.98$\\
% \rowcolor{green!10}
Qwen2.5-VL~\cite{Qwen2.5-VL} & Qwen2.5~\cite{yang2024qwen2-5} & 7B & $69.34$ & $65.43$ &$56.56$ & $24.56$ & $86.15$ & $81.61$\\
\hline
% \rowcolor{yellow!10}
\model~(w/o MoPE) & Qwen2.5 & 7B & $94.03$ & $93.20$ & $89.78$ &  $80.70$ & $98.27$ & $97.05$\\
% \rowcolor{yellow!10}
Qwen2.5-VL~(w MoPE) & Qwen2.5 & 7B & $75.04$ & $73.63$ & $59.23$ &  $31.58$ & $82.34$ & $87.69$\\
% \rowcolor{yellow!10}
\model & Qwen2.5 & 7B & $\textbf{94.05}$ & $\textbf{93.23}$ & $\textbf{89.78}$ & $\textbf{80.70}$ &  $\textbf{98.27}$ &$\textbf{97.09}$\\

\bottomrule
\end{tabular}
}
\caption{Comparison of \model~with other closed-source and Open-sourced VLMs on \benchqa.
All metrics in the table use Accuracy.
}
\label{tab:exp:result:qa}
\end{table*}
\begin{table*}[htbp]
\centering
\scalebox{0.8}{
\begin{tabular}{c|c|c|ccccccc}
\toprule
\multirow{2}{*}{\textbf{Model}} & \multicolumn{1}{c}{\textbf{MVBench}} & \multicolumn{1}{c}{\textbf{FAVOR}} & \multicolumn{7}{c}{\textbf{MotionBench}}    \\
\cline{2-10}
& ALL & ALL & ALL & MR & LM & CM & MO & AO & RC  \\
\hline
QWen2.5-VL-7B  & $68.94$ & $40.76$ & $46.23$ & $\textbf{45.11}$ &  $45.42$ & $\textbf{42.28}$ & $66.70$ & $\textbf{36.21}$ & $\textbf{33.08}$  \\
\model~(w/o \moext)  & $\textbf{69.03}$ & $\textbf{41.25}$ & $\textbf{47.41}$ & $45.09$ &  $\textbf{46.41}$ & $42.07$ & $\textbf{67.05}$ & $35.22$ & $32.98$\\
\model & $69.01$ & $41.02$ & $46.57$ & $45.06$ &  $46.33$ & $42.05$ & $67.02$ & $35.15$ & $32.90$\\

\bottomrule
\end{tabular}
}
\caption{The results of KPM-series on other motion-related benchmarks.}
\label{tab:exp:result:otherbench}
\end{table*}
\subsection{Metrics}

Following previous research about video caption and the traditions of the NLG (Natural Language Generation) fields~\cite{chen2024motionllm, lin-etal-2024-video, li2024videochat, maaz-etal-2024-video}, we use non-deteministic metrics such as GPT-Score, GPT-Accuarcy and Bert-Score~\cite{zhangbertscore} and use deterministic metrics such as BLEU~\cite{papineni-etal-2002-bleu}, Rouge\cite{lin-2004-rouge}, etc.
These metrics comprehensively evaluate the caption from multiple aspects, including grammar, semantics, the level of detail in describing the physical motion process, and the accuracy of the description.
For the QA task, we follow established practices from prior work~\cite{chen2024motionllm} and adopt accuracy based on option matching across different sub-tasks as the evaluation metric.
For hallucination evaluation, we employ two complementary metrics: GPT-Hall, which leverages GPT-4.1 as VLM judge to measure the average number of hallucinations in the caption, and Mo-Hall, which conducts precise evaluation based on the proposed \moext~ algorithm following Formula (9).
The design details of Mo-Hall and prompts used for VLMs are provided in Section C of Appendix.

\subsection{Implementation Details}
We perform full-parameter supervised fine-tuning (SFT) on the Qwen2.5-VL-7B-Instruct, with a learning rate set to $4 \times e^{-5}$. 
In the data construction and evaluation pipeline, we employ RTMPose3D~\cite{jiang2023rtmpose} for pose estimation, use xfm-bart-base parser from amrlib for AMR parsing, and apply SpaCy for dependency parsing.
The other details about implementation are described in Section D of Appendix.

\subsection{Baselines}

We compare the performance on the proposed \bench~ of the following baseline models including closed-source VLM APIs and open-source VLMs: GPT-4.1~\cite{gpt-4-1}, Gemini-2.5Pro~\cite{genimi-2p5}, Qwen2-VL~\cite{Qwen2VL}, Qwen2.5-VL~\cite{Qwen2.5-VL}, LLaVA-OneVision~\cite{li2024llavaov}, InternVideo-2.5~\cite{wang2025internvideo2}, MiniCPM-V2.6~\cite{yao2024minicpm}, VideoLLaMA3~\cite{zhang2025videollama}, and Tarsier2-Recap~\cite{yuan2025tarsier2}.
The details of these baseline are introduced in Section D of the Appendix.
We compare the performance of the following baseline models on the proposed \bench:

\subsection{Other Benchmarks}
To comprehensively assess both the quality of our dataset and the effectiveness of our model, we conduct evaluations on MVBench~\cite{li2024mvbench}, a widely used benchmark for evaluating general model capabilities, as well as on MotionBench~\cite{hong2024motionbench} and FAVOR~\cite{tu2025favor}, which place greater emphasis on fine-grained motion understanding.

\subsection{Results}
As shown in Table~\ref{tab:exp:result:main}, our \model~is significantly ahead in both ``content quality" and ``task accuracy", while maintaining an acceptable level of hallucination. 
Specifically, KPM (w/o MoPE) achieved higher scores in traditional NLG metrics, which are about 5-6 percentage points higher than the closed-source GPT-4.1, and also far ahead of the strongest open-source baseline Tarsier2-Recap.
Especially, the comparative advantage in GPT-Score indicating that our constructed dataset effectively enables the model to acquire the ability—through supervised fine-tuning—to unfold complex physical motions into detailed procedural descriptions. 
In the dimension of hallucination, Tarsier2-Recap maintains the lowest hallucination record both on GPT-Hall
and Mo-Hall, but it is slightly lacking in the ability to carry out complex motion processes.
However, by comparing \model~(w/o \moext) and \model, we can observe that the MoPE-based reward can effectively alleviate the hallucinations in motion description while only causing a marginal degradation in general NLG metrics, which demonstrates that \moext achieves a favorable trade-off between linguistic quality and factual consistency in motion description.

As shown in Table~\ref{tab:exp:result:qa} of \benchqa, 
\model~leads significantly with an overall accuracy of $94.05$, which is higher than the strongest open-source baseline InternVideo-2.5 and closed-source API GPT-4.1. 
It is worth noting that Environment task is a static scene recognition task that all models are generally good at. 
And the Camera dimension is still the most challenging link overall, revealing that perspective and lens reasoning are still the direction that needs to be specifically optimized in the next step. 
Further comparison of the ablation results shows that the direct gain of MoPE rewards on tasks is limited, but it does not bring side effects while maintaining high accuracy, indicating that the model has fully absorbed the reward signal and will not overfit specific question types. 
Overall, KPM has surpassed closed-source systems in dynamic scene understanding and emotion recognition, but perspective consistency and lens semantics are still the common shortcomings of current open source VLMs.

In addition, the results in Table~\ref{tab:exp:result:otherbench} demonstrate that KPM have excellent performance on other widely used QA benchmarks such as MVBench, FAVOR, and MotionBench, further verifying its wide adaptability and strong generalization ability in motion understanding and fine-grained video question answering. 
The results show that KPM not only has significant effects on self-built datasets, but also has excellent migration capabilities and question answering performance in multiple mainstream benchmarks, highlighting its application potential in actual multi-task scenarios.

Besides, the results of other cases and ablation experiments are presented in Section E of the Appendix.

\section{Conclusion}
\label{sec:conclusion}
In this work, we address two key challenges that limit the practical deployment of video captioning models: (1) the lack of fine-grained capability to describe motion details within videos, and (2) the high incidence of hallucinations during caption generation. 
To tackle these issues, we propose \textbf{\model}, an automated annotation pipeline based on motion decomposition and kinematic-based computation, and use it to construct a novel motion-centric video captioning benchmark, \textbf{\bench}.
This benchmark encompasses three main types of data: (i) video captions that provide detailed accounts of limb-specific movements during complex actions, (ii) a suite of motion-related complex question–answer pairs, and (iii) a challenging test set specifically designed for hallucination evaluation.
In addition, to further mitigate hallucinations in video captioning, we introduce \textbf{\moext}, the first linguistically driven motion attribute extraction algorithm, which precisely identifies motion-specific information in generated descriptions. 
Based on this extraction mechanism, we design a dedicated metric for accurately measuring motion hallucinations and innovatively integrate MoPE into the GRPO training framework. 
Experimental results demonstrate that this training scheme effectively reduces hallucination rates in video caption generation.

{
    \small
    \bibliographystyle{ieeenat_fullname}
    \bibliography{main}

\begin{thebibliography}{46}
\providecommand{\natexlab}[1]{#1}
\providecommand{\url}[1]{\texttt{#1}}
\expandafter\ifx\csname urlstyle\endcsname\relax
  \providecommand{\doi}[1]{doi: #1}\else
  \providecommand{\doi}{doi: \begingroup \urlstyle{rm}\Url}\fi

\bibitem[gen(2025)]{genimi-2p5}
{Gemini 2.5 Pro}.
\newblock \emph{https://deepmind.google/models/gemini/pro}, 2025.

\bibitem[gpt(2025)]{gpt-4-1}
{GPT-4.1}.
\newblock \emph{https://openai.com/index/gpt-4-1}, 2025.

\bibitem[Bai et~al.(2023)Bai, Bai, Yang, Wang, Tan, Wang, Lin, Zhou, and Zhou]{Qwen-VL}
Jinze Bai, Shuai Bai, Shusheng Yang, Shijie Wang, Sinan Tan, Peng Wang, Junyang Lin, Chang Zhou, and Jingren Zhou.
\newblock {Qwen-VL: A Versatile Vision-Language Model for Understanding, Localization, Text Reading, and Beyond}.
\newblock \emph{arXiv preprint arXiv:2308.12966}, 2023.

\bibitem[Bai et~al.(2025)Bai, Chen, Liu, Wang, Ge, Song, Dang, Wang, Wang, Tang, Zhong, Zhu, Yang, Li, Wan, Wang, Ding, Fu, Xu, Ye, Zhang, Xie, Cheng, Zhang, Yang, Xu, and Lin]{Qwen2.5-VL}
Shuai Bai, Keqin Chen, Xuejing Liu, Jialin Wang, Wenbin Ge, Sibo Song, Kai Dang, Peng Wang, Shijie Wang, Jun Tang, Humen Zhong, Yuanzhi Zhu, Mingkun Yang, Zhaohai Li, Jianqiang Wan, Pengfei Wang, Wei Ding, Zheren Fu, Yiheng Xu, Jiabo Ye, Xi Zhang, Tianbao Xie, Zesen Cheng, Hang Zhang, Zhibo Yang, Haiyang Xu, and Junyang Lin.
\newblock {Qwen2.5-VL Technical Report}.
\newblock \emph{arXiv preprint arXiv:2502.13923}, 2025.

\bibitem[Banarescu et~al.(2013)Banarescu, Bonial, Cai, Georgescu, Griffitt, Hermjakob, Knight, Koehn, Palmer, and Schneider]{banarescu2013abstract}
Laura Banarescu, Claire Bonial, Shu Cai, Madalina Georgescu, Kira Griffitt, Ulf Hermjakob, Kevin Knight, Philipp Koehn, Martha Palmer, and Nathan Schneider.
\newblock Abstract meaning representation for sembanking.
\newblock In \emph{Proceedings of the 7th linguistic annotation workshop and interoperability with discourse}, pages 178--186, 2013.

\bibitem[Chen et~al.(2024)Chen, Lu, Zeng, Zhang, Wang, Zhang, and Zhang]{chen2024motionllm}
Ling-Hao Chen, Shunlin Lu, Ailing Zeng, Hao Zhang, Benyou Wang, Ruimao Zhang, and Lei Zhang.
\newblock Motionllm: Understanding human behaviors from human motions and videos.
\newblock \emph{arXiv preprint arXiv:2405.20340}, 2024.

\bibitem[Cui et~al.(2023)Cui, Zhou, Yang, Wu, Zhang, Zou, and Yao]{cui2023holisticanalysishallucinationgpt4vision}
Chenhang Cui, Yiyang Zhou, Xinyu Yang, Shirley Wu, Linjun Zhang, James Zou, and Huaxiu Yao.
\newblock Holistic analysis of hallucination in gpt-4v(ision): Bias and interference challenges, 2023.

\bibitem[Fillmore et~al.()]{fillmore2006frame}
Charles~J Fillmore et~al.
\newblock Frame semantics.

\bibitem[Fillmore et~al.(1968)]{fillmore1968case}
Charles~J Fillmore et~al.
\newblock The case for case, universals in linguistic theory.
\newblock \emph{E. Bach and RT Harms. London}, 1968.

\bibitem[Guan et~al.(2024)Guan, Liu, Wu, Xian, Li, Liu, Wang, Chen, Huang, Yacoob, et~al.]{guan2024hallusionbench}
Tianrui Guan, Fuxiao Liu, Xiyang Wu, Ruiqi Xian, Zongxia Li, Xiaoyu Liu, Xijun Wang, Lichang Chen, Furong Huang, Yaser Yacoob, et~al.
\newblock Hallusionbench: an advanced diagnostic suite for entangled language hallucination and visual illusion in large vision-language models.
\newblock In \emph{Proceedings of the IEEE/CVF Conference on Computer Vision and Pattern Recognition}, pages 14375--14385, 2024.

\bibitem[Hong et~al.(2024)Hong, Cheng, Yang, Wang, Wang, Gu, Huang, Dong, and Tang]{hong2024motionbench}
Wenyi Hong, Yean Cheng, Zhuoyi Yang, Weihan Wang, Lefan Wang, Xiaotao Gu, Shiyu Huang, Yuxiao Dong, and Jie Tang.
\newblock Motionbench: Benchmarking and improving fine-grained video motion understanding for vision language models, 2024.

\bibitem[Jiang et~al.(2023)Jiang, Lu, Zhang, Ma, Han, Lyu, Li, and Chen]{jiang2023rtmpose}
Tao Jiang, Peng Lu, Li Zhang, Ningsheng Ma, Rui Han, Chengqi Lyu, Yining Li, and Kai Chen.
\newblock Rtmpose: Real-time multi-person pose estimation based on mmpose.
\newblock \emph{arXiv preprint arXiv:2303.07399}, 2023.

\bibitem[Jin et~al.(2020)Jin, Xu, Xu, Wang, Liu, Qian, Ouyang, and Luo]{jin2020whole}
Sheng Jin, Lumin Xu, Jin Xu, Can Wang, Wentao Liu, Chen Qian, Wanli Ouyang, and Ping Luo.
\newblock Whole-body human pose estimation in the wild.
\newblock In \emph{European Conference on Computer Vision}, pages 196--214. Springer, 2020.

\bibitem[Kingsbury and Palmer(2002)]{kingsbury-palmer-2002-treebank}
Paul Kingsbury and Martha Palmer.
\newblock From {T}ree{B}ank to {P}rop{B}ank.
\newblock In \emph{Proceedings of the Third International Conference on Language Resources and Evaluation ({LREC}{'}02)}, Las Palmas, Canary Islands - Spain, 2002. European Language Resources Association (ELRA).

\bibitem[Li et~al.(2024{\natexlab{a}})Li, Zhang, Guo, Zhang, Li, Zhang, Zhang, Zhang, Li, Liu, et~al.]{li2024llavaov}
Bo Li, Yuanhan Zhang, Dong Guo, Renrui Zhang, Feng Li, Hao Zhang, Kaichen Zhang, Peiyuan Zhang, Yanwei Li, Ziwei Liu, et~al.
\newblock Llava-onevision: Easy visual task transfer.
\newblock \emph{arXiv preprint arXiv:2408.03326}, 2024{\natexlab{a}}.

\bibitem[Li et~al.(2024{\natexlab{b}})Li, He, Wang, Li, Wang, Luo, Wang, Wang, and Qiao]{li2024videochat}
KunChang Li, Yinan He, Yi Wang, Yizhuo Li, Wenhai Wang, Ping Luo, Yali Wang, Limin Wang, and Yu Qiao.
\newblock Videochat: Chat-centric video understanding, 2024{\natexlab{b}}.

\bibitem[Li et~al.(2024{\natexlab{c}})Li, Wang, He, Li, Wang, Liu, Wang, Xu, Chen, Luo, et~al.]{li2024mvbench}
Kunchang Li, Yali Wang, Yinan He, Yizhuo Li, Yi Wang, Yi Liu, Zun Wang, Jilan Xu, Guo Chen, Ping Luo, et~al.
\newblock Mvbench: A comprehensive multi-modal video understanding benchmark.
\newblock In \emph{Proceedings of the IEEE/CVF Conference on Computer Vision and Pattern Recognition}, pages 22195--22206, 2024{\natexlab{c}}.

\bibitem[Lin et~al.(2024)Lin, Ye, Zhu, Cui, Ning, Jin, and Yuan]{lin-etal-2024-video}
Bin Lin, Yang Ye, Bin Zhu, Jiaxi Cui, Munan Ning, Peng Jin, and Li Yuan.
\newblock Video-{LL}a{VA}: Learning united visual representation by alignment before projection.
\newblock In \emph{Proceedings of the 2024 Conference on Empirical Methods in Natural Language Processing}, pages 5971--5984, Miami, Florida, USA, 2024. Association for Computational Linguistics.

\bibitem[Lin(2004)]{lin-2004-rouge}
Chin-Yew Lin.
\newblock {ROUGE}: A package for automatic evaluation of summaries.
\newblock In \emph{Text Summarization Branches Out}, pages 74--81, Barcelona, Spain, 2004. Association for Computational Linguistics.

\bibitem[Lin et~al.(2014)Lin, Maire, Belongie, Hays, Perona, Ramanan, Doll{\'a}r, and Zitnick]{lin2014microsoft}
Tsung-Yi Lin, Michael Maire, Serge Belongie, James Hays, Pietro Perona, Deva Ramanan, Piotr Doll{\'a}r, and C~Lawrence Zitnick.
\newblock Microsoft coco: Common objects in context.
\newblock In \emph{Computer vision--ECCV 2014: 13th European conference, zurich, Switzerland, September 6-12, 2014, proceedings, part v 13}, pages 740--755. Springer, 2014.

\bibitem[Liu et~al.(2023)Liu, Lin, Li, Wang, Yacoob, and Wang]{liu2023mitigating}
Fuxiao Liu, Kevin Lin, Linjie Li, Jianfeng Wang, Yaser Yacoob, and Lijuan Wang.
\newblock Mitigating hallucination in large multi-modal models via robust instruction tuning.
\newblock \emph{arXiv preprint arXiv:2306.14565}, 2023.

\bibitem[Maaz et~al.(2024)Maaz, Rasheed, Khan, and Khan]{maaz-etal-2024-video}
Muhammad Maaz, Hanoona Rasheed, Salman Khan, and Fahad Khan.
\newblock Video-{C}hat{GPT}: Towards detailed video understanding via large vision and language models.
\newblock In \emph{Proceedings of the 62nd Annual Meeting of the Association for Computational Linguistics (Volume 1: Long Papers)}, pages 12585--12602, Bangkok, Thailand, 2024. Association for Computational Linguistics.

\bibitem[McClellen and Rader(1979)]{dft1979}
Joseph~H McClellen and Charles~M Rader.
\newblock \emph{Number theory in digital signal processing}.
\newblock Prentice Hall Professional Technical Reference, 1979.

\bibitem[Mel'cuk et~al.(1988)]{mel1988dependency}
Igor~Aleksandrovic Mel'cuk et~al.
\newblock \emph{Dependency syntax: theory and practice}.
\newblock SUNY press, 1988.

\bibitem[Murray et~al.(2017)Murray, Li, and Sastry]{murray2017mathematical}
Richard~M Murray, Zexiang Li, and S~Shankar Sastry.
\newblock \emph{A mathematical introduction to robotic manipulation}.
\newblock CRC press, 2017.

\bibitem[Papineni et~al.(2002)Papineni, Roukos, Ward, and Zhu]{papineni-etal-2002-bleu}
Kishore Papineni, Salim Roukos, Todd Ward, and Wei-Jing Zhu.
\newblock {B}leu: a method for automatic evaluation of machine translation.
\newblock In \emph{Proceedings of the 40th Annual Meeting of the Association for Computational Linguistics}, pages 311--318, Philadelphia, Pennsylvania, USA, 2002. Association for Computational Linguistics.

\bibitem[Rodrigues(1840)]{rodrigues1840lois}
Olinde Rodrigues.
\newblock Des lois g{\'e}om{\'e}triques qui r{\'e}gissent les d{\'e}placements d'un syst{\`e}me solide dans l'espace, et de la variation des coordonn{\'e}es provenant de ces d{\'e}placements consid{\'e}r{\'e}s ind{\'e}pendamment des causes qui peuvent les produire.
\newblock \emph{Journal de math{\'e}matiques pures et appliqu{\'e}es}, 5:\penalty0 380--440, 1840.

\bibitem[Rohrbach et~al.(2018)Rohrbach, Hendricks, Burns, Darrell, and Saenko]{rohrbach2018object}
Anna Rohrbach, Lisa~Anne Hendricks, Kaylee Burns, Trevor Darrell, and Kate Saenko.
\newblock Object hallucination in image captioning.
\newblock \emph{arXiv preprint arXiv:1809.02156}, 2018.

\bibitem[Shao et~al.(2024)Shao, Wang, Zhu, Xu, Song, Bi, Zhang, Zhang, Li, et~al.]{shao2024deepseekmath}
Zhihong Shao, Peiyi Wang, Qihao Zhu, Runxin Xu, Junxiao Song, Xiao Bi, Haowei Zhang, Mingchuan Zhang, YK Li, et~al.
\newblock Deepseekmath: Pushing the limits of mathematical reasoning in open language models.
\newblock \emph{arXiv preprint arXiv:2402.03300}, 2024.

\bibitem[Talmy(1985)]{talmy1985lexicalization}
Leonard Talmy.
\newblock Lexicalization patterns: Semantic structure in lexical forms.
\newblock \emph{Language typology and syntactic description}, 3\penalty0 (99):\penalty0 36--149, 1985.

\bibitem[Tu et~al.(2025)Tu, Zhang, Chen, Ye, Zeng, Cheng, Yu, and Chen]{tu2025favor}
Chongjun Tu, Lin Zhang, Pengtao Chen, Peng Ye, Xianfang Zeng, Wei Cheng, Gang Yu, and Tao Chen.
\newblock Favor-bench: A comprehensive benchmark for fine-grained video motion understanding.
\newblock \emph{arXiv preprint arXiv:2503.14935}, 2025.

\bibitem[Villegas et~al.(2023)Villegas, Babaeizadeh, Kindermans, Moraldo, Zhang, Saffar, Castro, Kunze, and Erhan]{villegas2023phenaki}
Ruben Villegas, Mohammad Babaeizadeh, Pieter-Jan Kindermans, Hernan Moraldo, Han Zhang, Mohammad~Taghi Saffar, Santiago Castro, Julius Kunze, and Dumitru Erhan.
\newblock Phenaki: Variable length video generation from open domain textual descriptions.
\newblock In \emph{ICLR}, 2023.

\bibitem[Wang et~al.(2023{\natexlab{a}})Wang, Zhou, Xu, Shi, Zhao, Xu, Ye, Yan, Zhang, Zhu, et~al.]{wang2023evaluation}
Junyang Wang, Yiyang Zhou, Guohai Xu, Pengcheng Shi, Chenlin Zhao, Haiyang Xu, Qinghao Ye, Ming Yan, Ji Zhang, Jihua Zhu, et~al.
\newblock Evaluation and analysis of hallucination in large vision-language models.
\newblock \emph{arXiv preprint arXiv:2308.15126}, 2023{\natexlab{a}}.

\bibitem[Wang et~al.(2024)Wang, Bai, Tan, Wang, Fan, Bai, Chen, Liu, Wang, Ge, Fan, Dang, Du, Ren, Men, Liu, Zhou, Zhou, and Lin]{Qwen2VL}
Peng Wang, Shuai Bai, Sinan Tan, Shijie Wang, Zhihao Fan, Jinze Bai, Keqin Chen, Xuejing Liu, Jialin Wang, Wenbin Ge, Yang Fan, Kai Dang, Mengfei Du, Xuancheng Ren, Rui Men, Dayiheng Liu, Chang Zhou, Jingren Zhou, and Junyang Lin.
\newblock {Qwen2-VL: Enhancing Vision-Language Model's Perception of the World at Any Resolution}.
\newblock \emph{arXiv preprint arXiv:2409.12191}, 2024.

\bibitem[Wang et~al.(2023{\natexlab{b}})Wang, Yuan, Zhang, Chen, Wang, Zhang, Shen, Zhao, and Zhou]{wang2023videocomposer}
Xiang Wang, Hangjie Yuan, Shiwei Zhang, Dayou Chen, Jiuniu Wang, Yingya Zhang, Yujun Shen, Deli Zhao, and Jingren Zhou.
\newblock Videocomposer: Compositional video synthesis with motion controllability.
\newblock \emph{Advances in Neural Information Processing Systems}, 36:\penalty0 7594--7611, 2023{\natexlab{b}}.

\bibitem[Wang et~al.(2025)Wang, Li, Yan, He, Yu, Zeng, Wang, Ma, Huang, Gao, et~al.]{wang2025internvideo2}
Yi Wang, Xinhao Li, Ziang Yan, Yinan He, Jiashuo Yu, Xiangyu Zeng, Chenting Wang, Changlian Ma, Haian Huang, Jianfei Gao, et~al.
\newblock Internvideo2.5: Empowering video mllms with long and rich context modeling.
\newblock \emph{arXiv preprint arXiv:2501.12386}, 2025.

\bibitem[Wu et~al.(2024)Wu, Huang, Zhou, Ying, Wang, Lin, and Chen]{wu2024internlm2-5}
Zijian Wu, Suozhi Huang, Zhejian Zhou, Huaiyuan Ying, Jiayu Wang, Dahua Lin, and Kai Chen.
\newblock Internlm2. 5-stepprover: Advancing automated theorem proving via expert iteration on large-scale lean problems.
\newblock \emph{arXiv preprint arXiv:2410.15700}, 2024.

\bibitem[Xu et~al.(2022)Xu, Jin, Liu, Qian, Ouyang, Luo, and Wang]{xu2022zoomnas}
Lumin Xu, Sheng Jin, Wentao Liu, Chen Qian, Wanli Ouyang, Ping Luo, and Xiaogang Wang.
\newblock Zoomnas: searching for whole-body human pose estimation in the wild.
\newblock \emph{IEEE Transactions on Pattern Analysis and Machine Intelligence}, 45\penalty0 (4):\penalty0 5296--5313, 2022.

\bibitem[Yang et~al.(2024{\natexlab{a}})Yang, Yang, Hui, Zheng, Yu, Zhou, Li, Li, Liu, Huang, Dong, Wei, Lin, Tang, Wang, Yang, Tu, Zhang, Ma, Yang, Xu, Zhou, Bai, He, Lin, Dang, Lu, Chen, Yang, Li, Xue, Ni, Zhang, Wang, Peng, Men, Gao, Lin, Wang, Bai, Tan, Zhu, Li, Liu, Ge, Deng, Zhou, Ren, Zhang, Wei, Ren, Liu, Fan, Yao, Zhang, Wan, Chu, Liu, Cui, Zhang, Guo, and Fan]{yang2024qwen2technicalreport}
An Yang, Baosong Yang, Binyuan Hui, Bo Zheng, Bowen Yu, Chang Zhou, Chengpeng Li, Chengyuan Li, Dayiheng Liu, Fei Huang, Guanting Dong, Haoran Wei, Huan Lin, Jialong Tang, Jialin Wang, Jian Yang, Jianhong Tu, Jianwei Zhang, Jianxin Ma, Jianxin Yang, Jin Xu, Jingren Zhou, Jinze Bai, Jinzheng He, Junyang Lin, Kai Dang, Keming Lu, Keqin Chen, Kexin Yang, Mei Li, Mingfeng Xue, Na Ni, Pei Zhang, Peng Wang, Ru Peng, Rui Men, Ruize Gao, Runji Lin, Shijie Wang, Shuai Bai, Sinan Tan, Tianhang Zhu, Tianhao Li, Tianyu Liu, Wenbin Ge, Xiaodong Deng, Xiaohuan Zhou, Xingzhang Ren, Xinyu Zhang, Xipin Wei, Xuancheng Ren, Xuejing Liu, Yang Fan, Yang Yao, Yichang Zhang, Yu Wan, Yunfei Chu, Yuqiong Liu, Zeyu Cui, Zhenru Zhang, Zhifang Guo, and Zhihao Fan.
\newblock Qwen2 technical report, 2024{\natexlab{a}}.

\bibitem[Yang et~al.(2024{\natexlab{b}})Yang, Yang, Zhang, Hui, Zheng, Yu, Li, Liu, Huang, Wei, et~al.]{yang2024qwen2-5}
An Yang, Baosong Yang, Beichen Zhang, Binyuan Hui, Bo Zheng, Bowen Yu, Chengyuan Li, Dayiheng Liu, Fei Huang, Haoran Wei, et~al.
\newblock Qwen2. 5 technical report.
\newblock \emph{arXiv preprint arXiv:2412.15115}, 2024{\natexlab{b}}.

\bibitem[Yao et~al.(2024)Yao, Yu, Zhang, Wang, Cui, Zhu, Cai, Li, Zhao, He, et~al.]{yao2024minicpm}
Yuan Yao, Tianyu Yu, Ao Zhang, Chongyi Wang, Junbo Cui, Hongji Zhu, Tianchi Cai, Haoyu Li, Weilin Zhao, Zhihui He, et~al.
\newblock Minicpm-v: A gpt-4v level mllm on your phone.
\newblock \emph{arXiv preprint arXiv:2408.01800}, 2024.

\bibitem[Yuan et~al.(2025)Yuan, Wang, Sun, Zhang, and Lin]{yuan2025tarsier2}
Liping Yuan, Jiawei Wang, Haomiao Sun, Yuchen Zhang, and Yuan Lin.
\newblock Tarsier2: Advancing large vision-language models from detailed video description to comprehensive video understanding.
\newblock \emph{arXiv preprint arXiv:2501.07888}, 2025.

\bibitem[Zhai et~al.(2023)Zhai, Yang, Zhao, Xu, Shen, Zhao, Keutzer, Li, Yan, and Fan]{zhai2023halle}
Bohan Zhai, Shijia Yang, Xiangchen Zhao, Chenfeng Xu, Sheng Shen, Dongdi Zhao, Kurt Keutzer, Manling Li, Tan Yan, and Xiangjun Fan.
\newblock Halle-switch: Rethinking and controlling object existence hallucinations in large vision-language models for detailed caption.
\newblock 2023.

\bibitem[Zhang et~al.(2025)Zhang, Li, Cheng, Hu, Yuan, Chen, Leng, Jiang, Zhang, Li, et~al.]{zhang2025videollama}
Boqiang Zhang, Kehan Li, Zesen Cheng, Zhiqiang Hu, Yuqian Yuan, Guanzheng Chen, Sicong Leng, Yuming Jiang, Hang Zhang, Xin Li, et~al.
\newblock Videollama 3: Frontier multimodal foundation models for image and video understanding.
\newblock \emph{arXiv preprint arXiv:2501.13106}, 2025.

\bibitem[Zhang et~al.(2020)Zhang, Kishore, Wu, Weinberger, and Artzi]{zhangbertscore}
Tianyi Zhang, Varsha Kishore, Felix Wu, Kilian~Q Weinberger, and Yoav Artzi.
\newblock Bertscore: Evaluating text generation with bert.
\newblock In \emph{International Conference on Learning Representations}, 2020.

\bibitem[Zhu et~al.(2023)Zhu, Ma, Liu, Liu, Wu, and Wang]{motionbert}
Wentao Zhu, Xiaoxuan Ma, Zhaoyang Liu, Libin Liu, Wayne Wu, and Yizhou Wang.
\newblock Motionbert: A unified perspective on learning human motion representations.
\newblock In \emph{Proceedings of the IEEE/CVF International Conference on Computer Vision (ICCV)}, pages 15085--15099, 2023.

\end{thebibliography}
}

\clearpage
% \setcounter{page}{1}
% \maketitlesupplementary
% \maketitleappendix
\appendix

\begin{figure*}[t]
\centering
\includegraphics[width=1.0\linewidth]{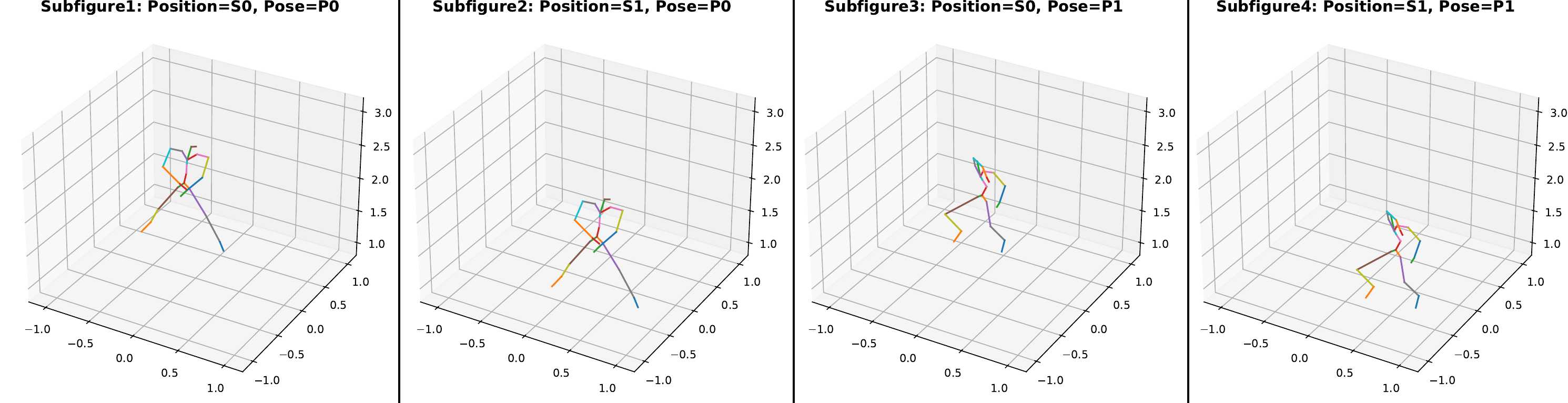}
\caption{
The human motion~(subfigure $1$ $\rightarrow$ subfigure $4$) can be decomposing into Position Translation~(subfigure $1$ $\rightarrow$ subfigure $2$) and Postural Transformation~(subfigure $1$ $\rightarrow$ subfigure $3$).
}
\label{fig:motion:physical:example1}
\end{figure*}
\section{Motion Decomposition}
In Section 1 of the main paper, we propose that human motion can be decomposed into \textbf{position translation} and \textbf{postural transformation}. Figure~\ref{fig:motion:physical:example1} provides an illustrative depiction of this decomposition process.

The motion trajectory of the person in the figure can be interpreted as a transition from the initial state $(s_0, p_0)$ in Subfigure 1 to the final state $(s_1, p_1)$ in Subfigure 4. 
This process can be viewed as the superposition of two sub-motions.

In the first sub-motion, the person moves from $(s_0, p_0)$ to $(s_1, p_0)$, keeping the posture $p_0$ unchanged while translating the position from $s_0$ to $s_1$.

In the second sub-motion, the person transitions from $(s_0, p_0)$ to $(s_0, p_1)$, maintaining a fixed position $s_0$ while changing posture from $p_0$ to $p_1$.

According to basic vector decomposition and composition principles, the original motion $(s_0, p_0) \rightarrow (s_1, p_1)$ can be regarded as the combination of these two sub-motions.

\section{Details of KPM Pipeline}
\subsection{The Calculable Joint Angles}

As described in Section 3 of the main paper, we adopt RTMPose3D as our 3D pose estimation model, which utilizes the COCO-Wholebody-133 skeletal representation.
Based on fundamental kinematic principles, we construct 10 computable and motion-relevant joint angles from the 133 skeletal keypoints to capture meaningful aspects of body posture and dynamics.
In Table~\ref{tab:joint_angles_schemaA}, we provide the details of the 10 selected joints used for angular velocity computation, including the specific triplets of skeletal keypoints that define each joint angle.

\begin{figure*}[t]
\centering
\includegraphics[width=1.0\linewidth]{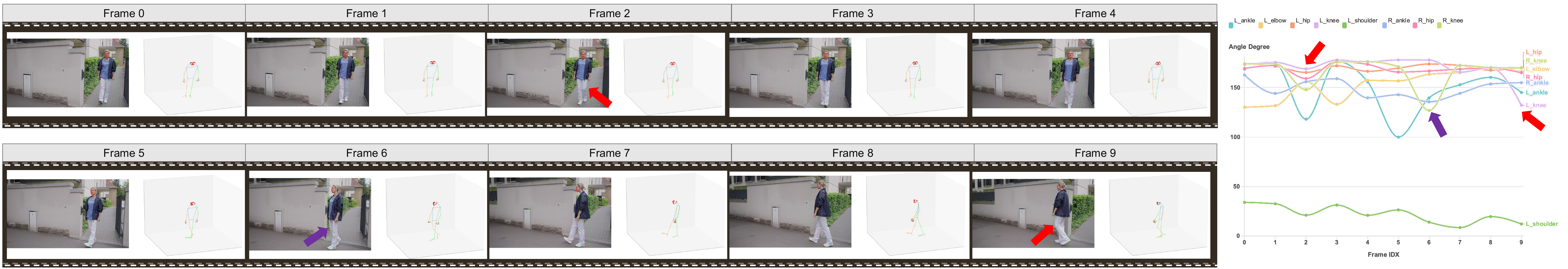}
\caption{
The case of the joint angles calculation.
}
\label{fig:phy:calc:case}
\end{figure*}

Additionally, the case of the kinematic-based computation is illustrated in Figure~\ref{fig:phy:calc:case}. 
In the left panel, it can be observe that the subject’s right knee angle exhibits a pronounced flexion at the 2nd and 9th frames compared with the preceding frames, as indicated by the red arrows. 
Correspondingly, in the right panel, the curve labeled right\_knee shows clear sharp changes at the same frames. 
Similarly, the left panel shows a noticeable flexion of the subject’s left knee at the 6th frame, highlighted by the purple arrow, which aligns with a marked change in the left\_knee curve at the corresponding frame in the right panel.
This alignment between visual observation and computed motion data provides strong validation for the correctness of our kinematic-based calculations.

\begin{table*}[t]
\centering
\scalebox{0.6}{
\begin{tabular}{@{}lll@{}}
\toprule
\textbf{Joint Angle}          & \textbf{Joint Point Combination (index, name)}                                  & \textbf{Description}                                                \\ \midrule
Left Shoulder (LShoulder)     & (7, 5, 11) – Left Elbow, Left Shoulder, Left Hip                                & Position of the left upper arm relative to the torso.               \\
Right Shoulder (RShoulder)    & (8, 6, 12) – Right Elbow, Right Shoulder, Right Hip                             & Position of the right upper arm relative to the torso.              \\
Left Elbow (LElbow)           & (5, 7, 9) – Left Shoulder, Left Elbow, Left Wrist                               & Flexion/extension of the left elbow (forearm relative to upper arm). \\
Right Elbow (RElbow)          & (6, 8, 10) – Right Shoulder, Right Elbow, Right Wrist                           & Flexion/extension of the right elbow (forearm relative to upper arm). \\
Left Hip (LHip)               & (5, 11, 13) – Left Shoulder, Left Hip, Left Knee                                & Angle between the left thigh and the torso.                         \\
Right Hip (RHip)              & (6, 12, 14) – Right Shoulder, Right Hip, Right Knee                             & Angle between the right thigh and the torso.                        \\
Left Knee (LKnee)             & (11, 13, 15) – Left Hip, Left Knee, Left Ankle                                  & Flexion/extension of the left knee (lower leg relative to thigh).   \\
Right Knee (RKnee)            & (12, 14, 16) – Right Hip, Right Knee, Right Ankle                               & Flexion/extension of the right knee (lower leg relative to thigh).  \\
Left Ankle (LAnkle)           & (13, 15, 19 / 17) – Left Knee, Left Ankle, Left Heel / Left Big Toe             & Dorsiflexion/plantarflexion of the left ankle (foot relative to lower leg; heel preferred, big toe as fallback). \\
Right Ankle (RAnkle)          & (14, 16, 22 / 20) – Right Knee, Right Ankle, Right Heel / Right Big Toe         & Dorsiflexion/plantarflexion of the right ankle (foot relative to lower leg; heel preferred, big toe as fallback). \\ \bottomrule
\end{tabular}
}
\caption{Definition of the 10 joint hinge angles used in KPM pipeline. Each angle is computed at the middle joint from its two adjacent segments.}
\label{tab:joint_angles_schemaA}
\end{table*}

% \lin{insert MoPE case}
\begin{algorithm}[t]
\caption{\textsc{Extract Motion Attributes And Sequence}}
\KwIn{Raw caption text $x$}
\KwOut{Ordered list of structured actions $A$}

\textbf{Preprocessing:} \\
Resolve coreference in $x$; obtain dependency parse $D$ and AMR graph string $S$. \\
\If{$S$ invalid}{
    \Return empty list
}
Decode $S$ into AMR graph $G$.

\textbf{Action Identification:} \\
Extract AMR action candidates $C$ from $G$, filtering stative concepts. \\
Initialize empty combined action list $A$ and mapping $M$.

\ForEach{AMR node $c_i \in C$}{
    Initialize action entry $a_i$. \\
    Find matching verb token $t_i$ in $D$ via lemma alignment and validity filter. \\
    \If{$t_i$ found}{
        Populate $a_i$ with fused AMR and DP attributes (subject, object, direction, modifiers).
        Update $M[t_i] \gets c_i$.
    }
    \Else{
        Populate $a_i$ using AMR attributes only.
    }
    Add $a_i$ to $A$.
}

\textbf{Temporal Relation Extraction:} \\
Extract temporal links $L$ using a prioritized strategy:
\begin{enumerate}
    \item Inter-sentence connectives (Dependency Parsing)
    \item Intra-sentence connectives (Dependency Parsing)
    \item Control verb dependencies (AMR/Dependency Parsing)
    \item AMR time roles
    \item Implicit sequence (Dependency Parsing)
\end{enumerate}
Deduplicate and prioritize $L$.

\textbf{Action Sequencing:} \\
Topologically sort $A$ using $L$ (Kahn's algorithm). \\
Assign temporal order indices and next-action relations.

\Return $A$
\label{alg:post:parsing:extraction}
\end{algorithm}
\begin{figure*}[t]
\centering
\includegraphics[width=1.0\linewidth]{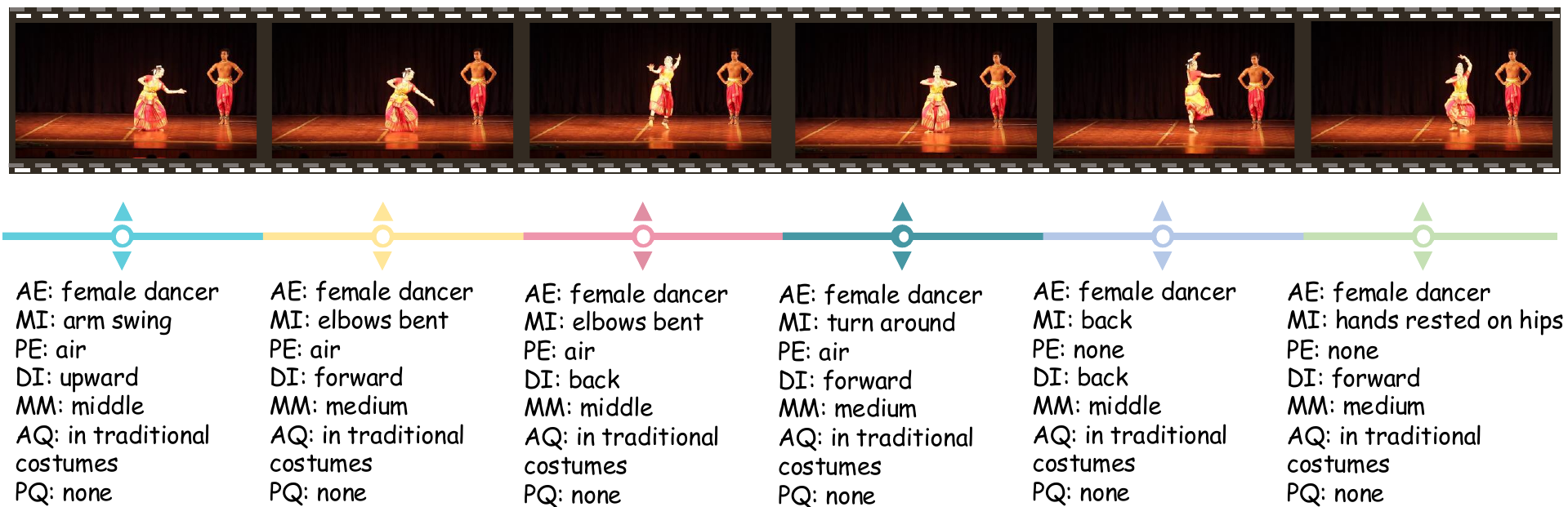}
\caption{
An example of PaMoR. The motion process description is decomposed into seven core elements: MI (Motion Initiation), AE (Agent Entity), PE (Patient Entity), MM (Magnitude Modifier), DI (Direction Indicator), AQ (Agent Qualifier), and PQ (Patient Qualifier). These elements are expressed using three types of units.
}
\label{fig:unit:case}
\end{figure*}

\subsection{Confidence Thresholding and Joint Validity in 3D Based Kinematics}
For each 3D pose position, the pose estimation model provides a confidence score.
We adopt a fixed threshold of 0.6 following the common visualization settings of pose estimation models to decide whether a joint is considered reliable, and we apply this threshold consistently for both visualization and kinematic attribute computation. 
Skeletons whose confidence falls below 0.6 are treated as invalid (i.e., missing) and excluded from subsequent calculations.
For joint angles, which are defined over triplets of skeletons, we impose a stricter criterion: an angle is computed only when all three constituent joints have confidence scores greater than 0.6. 
Otherwise, the corresponding angle is discarded and not used in any derived kinematic features. 
This filtering strategy reduces the influence of noisy pose estimates on downstream kinematic calculation.
In addition, we incorporate a robustness mechanism to handle degenerate cases. 
For a small fraction of videos, either no human is present at all, or all detected 3D pose estimation results are deemed unreliable under the above confidence rules. 
In these cases, our pipeline falls back to generating captions directly from the raw video content, without using any kinematic attributes. 
This fallback strategy prevents the model from producing spurious pose-based analyses when the underlying pose estimation fails.

\subsection{Prompts}
\begin{figure*}[t]
\centering
\includegraphics[width=1.0\linewidth]{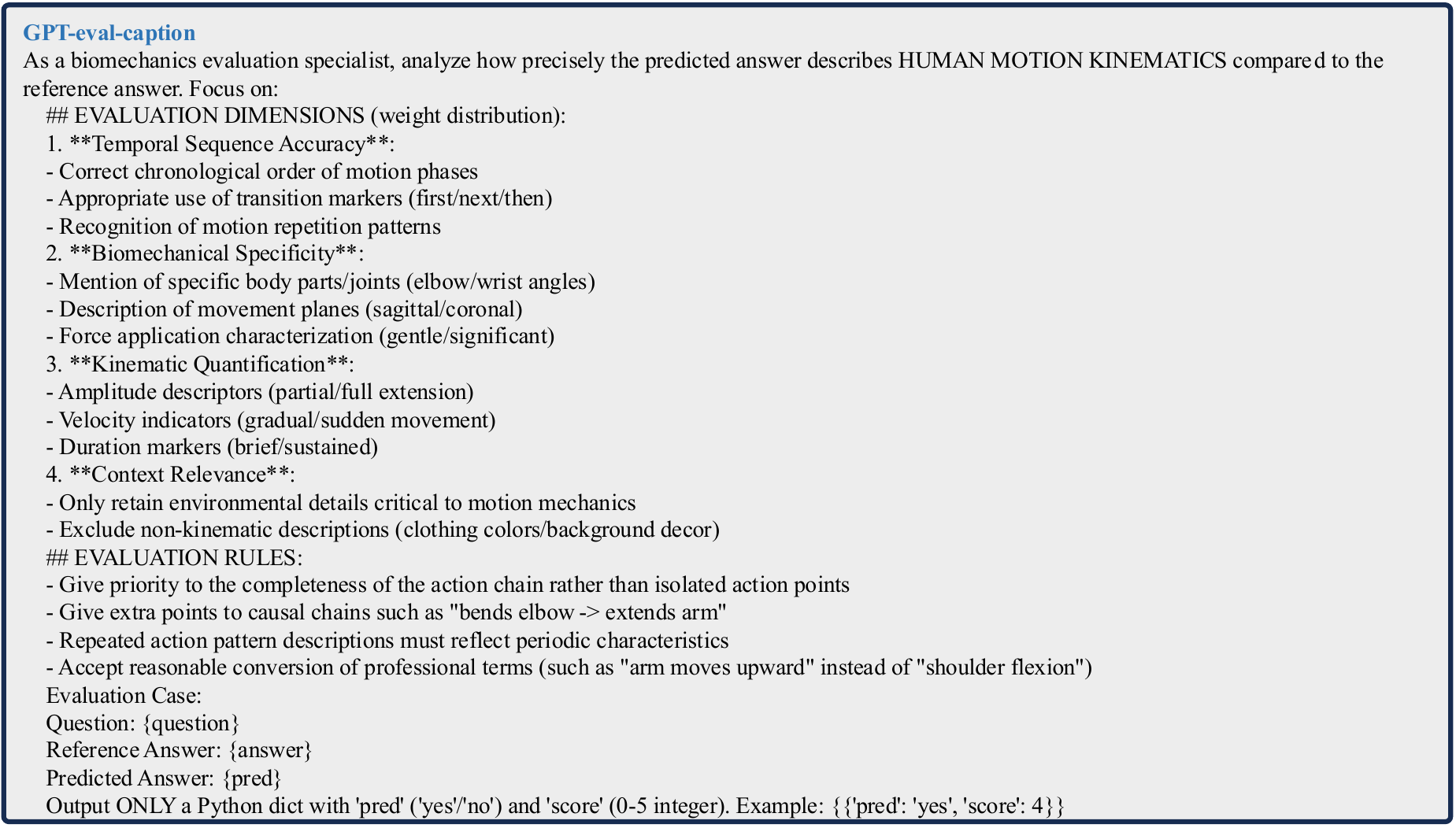}
\caption{
Prompt for evaluating caption quality.
}
\label{fig:prompt:gpt-eval-caption}
\end{figure*}

\begin{figure*}[t]
\centering
\includegraphics[width=0.8\linewidth]{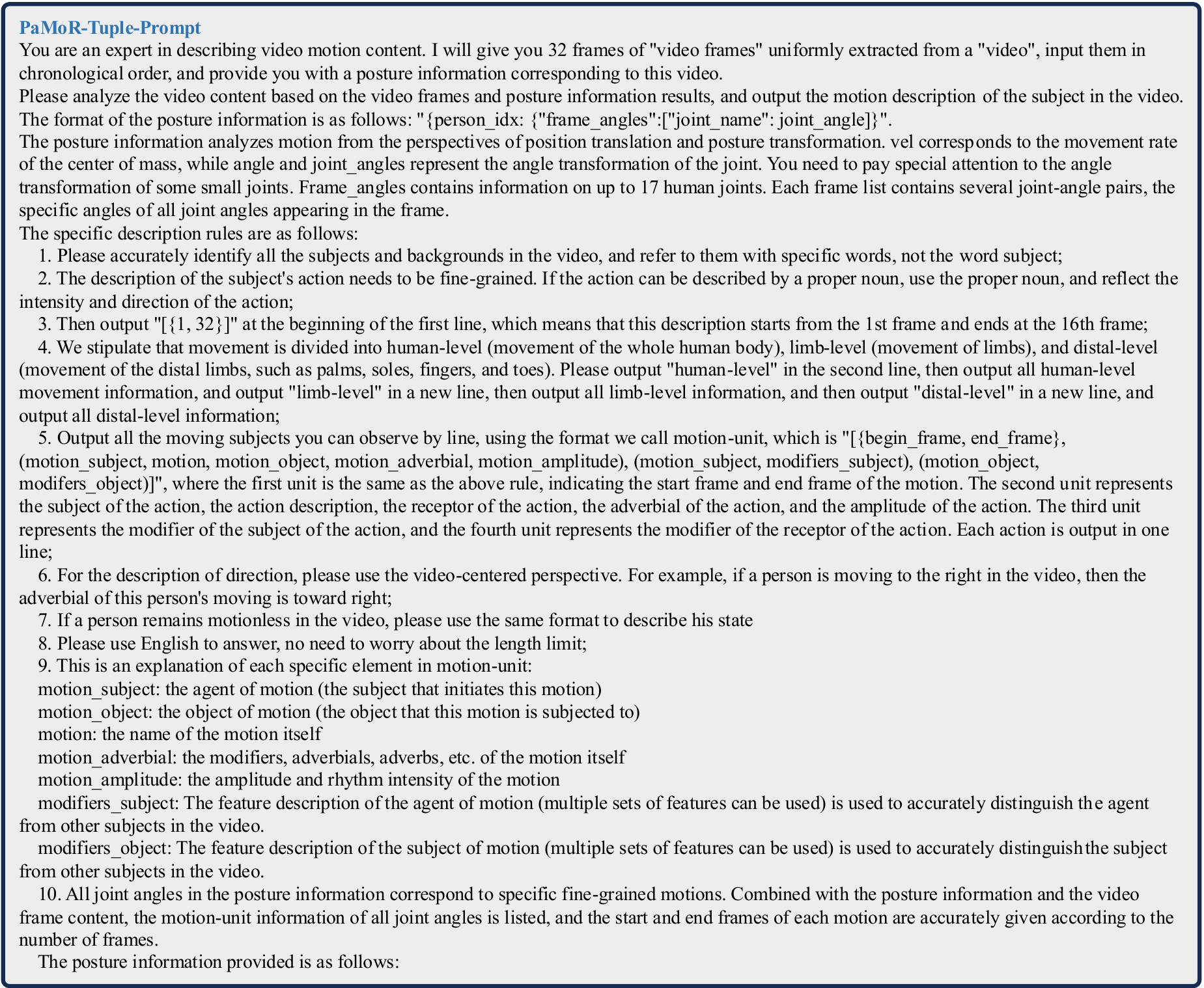}
\caption{
Prompt of PaMoR-Tuple.
}
\label{fig:prompt:pamor-physical-prompt}
\end{figure*}

\begin{figure*}[t]
\centering
\includegraphics[width=0.8\linewidth]{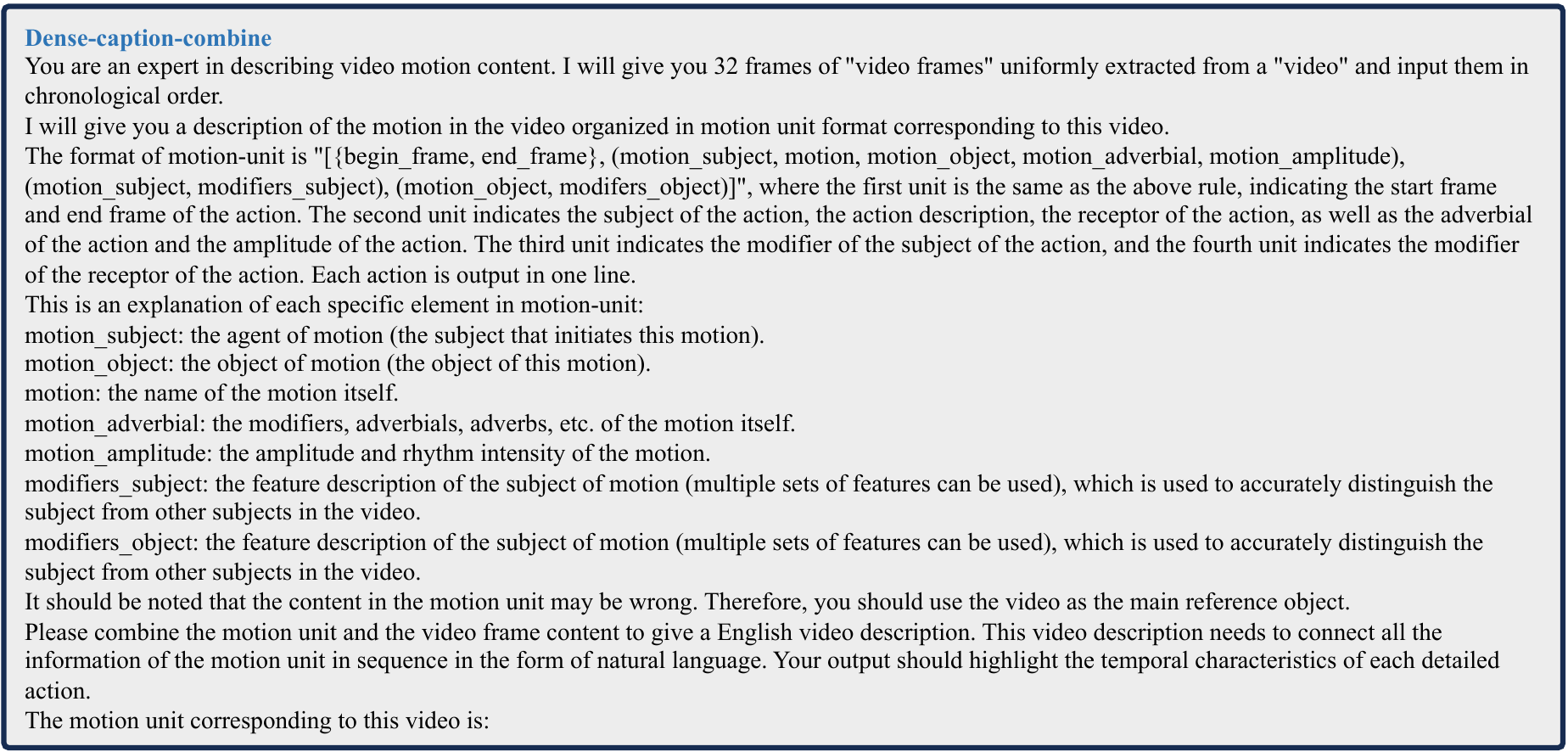}
\caption{
Prompt of PaMoR-Caption.
}
\label{fig:prompt:dense-caption-combine}
\end{figure*}

In this section, we list the prompts used in KPM pipeline. 
Since the capabilities of GPT are widely recognized by the community, we also measure the metrics by using GPT-4.1 as a judge. 
In Figure~\ref{fig:prompt:gpt-eval-caption}, we show the prompts used to measure the quality of the caption.
In Figure~\ref{fig:prompt:pamor-physical-prompt} and Figure~\ref{fig:prompt:dense-caption-combine}, we show the prompt used to generate a PaMoR-tuple with a prompt by injecting kinematic information, and the prompt used to combine the PaMoR-tuple into the final caption.

\begin{figure*}[t]
\centering
\includegraphics[width=0.8\linewidth]{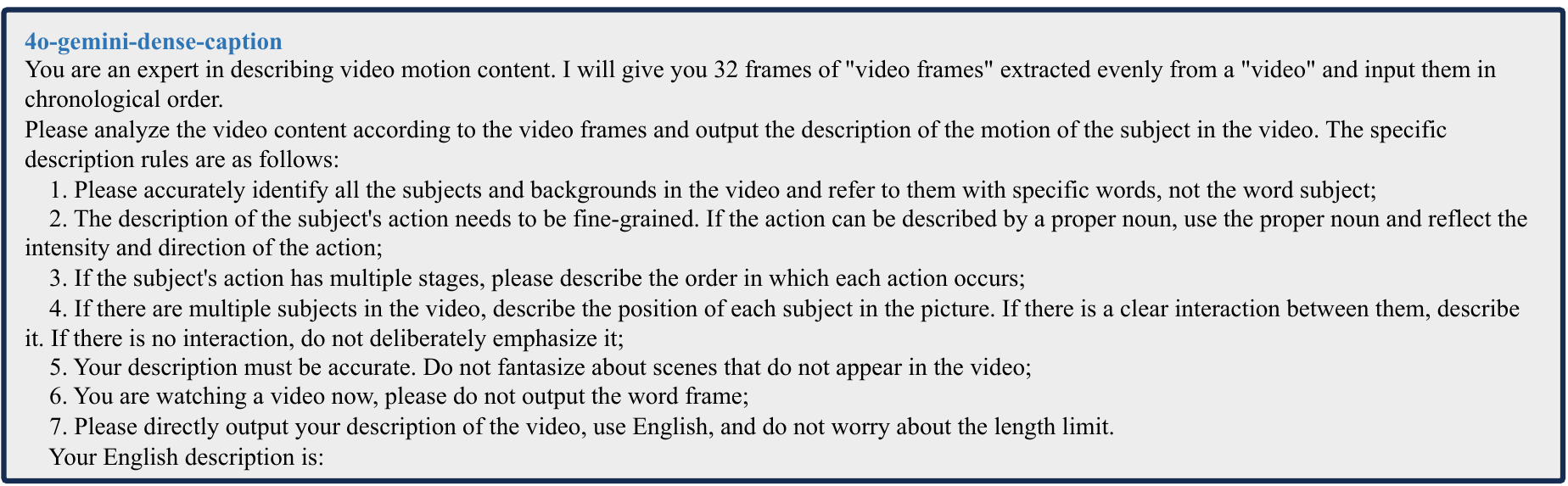}
\caption{
Prompt for generating dense caption from GPT and Gemini.
}
\label{fig:prompt:4o-gemini-dense-caption}
\end{figure*}

In order to fairly compare the capabilities of other models, we carefully designed prompts to call GPT-4.1 or Gemini-2.5-Pro to generate high-quality captions, as shown in Figure~\ref{fig:prompt:4o-gemini-dense-caption}. 
This way, the evaluation results of the captions they generate will be convincing when compared with ours. 
Otherwise, it is difficult to stimulate GPT-4.1 and Gemini-2.5-Pro's capabilities by giving them a simple prompt to generate captions.

In designing the evaluation prompts, we follow the approach established in prior research, assessing the quality of generated captions from multiple perspectives such as correctness and content alignment. 
Besides, we also add the evaluation to the ability to unfold motion processes, allowing us to evaluate whether the model can accurately elaborate the dynamics of complex actions.

\subsection{Algorithm}
The pseudo-code of the \moext~algorithm is shown in Algorithm 1.

Given a caption $x$, our goal is to recover an ordered sequence of structured actions
$A = {a_1, \dots, a_n}$, where each $a_i$ contains a motion predicate, its agent and patient, coarse directional information, and a temporal index. 
Algorithm 1 summarizes the overall pipeline. Below we describe the main components.

\subsubsection{Preprocessing and Graph Construction}

We first apply an off-the-shelf dependency parser to obtain sentence-level dependency trees $D$ for the caption. In parallel, we run an AMR parser to produce a graph string, which is then decoded into an AMR graph $G$.

\subsubsection{Action Candidate Extraction and Alignment}

\paragraph{AMR-based candidate extraction.}
We first identify action candidates purely from the AMR graph. 
For each instance node $(v, c)$ in $G$, we treat it as a potential action if its concept $c$ follows the PropBank style and ends with ``-01'' (e.g., \emph{walk-01}, \emph{turn-01}). 
For each remaining action variable $v$, we collect outgoing semantic roles, including \texttt{:ARG0}, \texttt{:ARG1}, \texttt{:ARG2}, \texttt{:direction}, \texttt{:manner}, \texttt{:location} and \texttt{:time}. 
These roles will be used to infer subjects, objects, directions and temporal cues.

\paragraph{Dependency alignment.}
To ground AMR actions in the surface text, we align each AMR predicate to a verb token in the dependency trees. 
Given a concept $c$ such as \emph{walk-01}, we extract its lemma (e.g., \emph{walk}) and search for tokens whose lemmatized form matches this lemma.

We then apply a dynamic-verb filter \texttt{is\_valid\_action\_verb}, which removes:
(i) auxiliaries and copulas (dependencies \texttt{aux}, \texttt{auxpass}, \texttt{cop});
(ii) verbs that only act as nominal modifiers (\texttt{acl}, \texttt{amod} attached to nouns or pronouns); and
(iii) most \texttt{advcl} verbs that lack an explicit temporal marker. For the last case, adverbial clause verbs are only kept when the clause contains a subordinating conjunction that is a known temporal connective (e.g., ``after'', ``before'', ``when'').

If exactly one token survives this filter, we align it to the AMR action; if multiple candidates exist, we use a simple heuristic and choose the earliest token in linear order; if none is found, the action remains AMR-only. For aligned actions, we record the verbal lemma (\texttt{action\_verb}) and its character span in the original caption.

\paragraph{Attribute fusion.}
For each action, we then fuse AMR roles and dependency relations into a single structured entry.

\emph{Subject.}
We first look for an AMR \texttt{:ARG0} role. If the corresponding node has a \texttt{:name} subgraph, we recover a natural name from the \texttt{:op1} literal; otherwise we use the concept label as the subject text. If AMR does not provide \texttt{:ARG0}, we fall back to the dependency tree, using \texttt{nsubj}/\texttt{nsubjpass} and a head-finding heuristic to obtain the subject phrase.

\emph{Object.}
We follow the same strategy for the patient: we prioritize AMR \texttt{:ARG1}, and only fall back to dependency objects (e.g., \texttt{dobj} or verb–preposition–object patterns) when AMR is missing or underspecified.

\emph{Direction and modifiers.}
We explicitly track a small lexicon of directional words (e.g., \emph{forward}, \emph{left}, \emph{right}, \emph{up}, \emph{down}). 
If AMR roles \texttt{:direction} or \texttt{:ARG2} point to a concept or literal in this lexicon, we treat it as the motion direction. 
Otherwise, we derive a direction from dependency-based adverbial or prepositional modifiers (e.g., \texttt{advmod} or \texttt{prep}–\texttt{pobj} such as ``to the right''). 
Remaining adverbs and spatial prepositional phrases are collected as general modifiers. 
From AMR we additionally include \texttt{:manner} and \texttt{:location} roles, marking locations with a \texttt{location:} prefix. 
The resulting modifier list is deduplicated and sorted.

Each action $a_i$ thus aggregates a predicate, subject, object, direction and other modifiers, together with its AMR variable ID and the span of the aligned verb token.

\subsubsection{Temporal Relation Extraction}

We next infer temporal relations between actions by combining clause-level dependency cues and AMR time roles.

\paragraph{Dependency-based cues.}
For each sentence, we collect all verbs that (i) pass \texttt{is\_valid\_action\_verb} and (ii) have been aligned to AMR actions, and sort them by token index. We then add edges of different types.

First, for adverbial clauses (\texttt{advcl}), if a clause verb has a child with dependency \texttt{mark} and its lemma is a temporal connective (e.g., ``after'', ``before'', ``when''), we connect the corresponding clause and main-clause actions with an explicit temporal edge. Connectives such as ``after'', ``since'', ``once'', ``following'' and ``upon'' indicate that the clause action is temporally later than the main action, while connectives such as ``before'', ``until'', ``when'' and ``while'' reverse this ordering.

Second, for conjunctions (\texttt{conj}), we look for temporal adverbs like ``then'' modifying the conjunct verb, or a coordinating conjunction \texttt{cc} equal to ``and'' between the head and conjunct verbs. The former case yields an explicit temporal edge; the latter is treated as an implicit sequential cue, typically indicating that the head action precedes the conjunct action.

Finally, when two adjacent verbs in the same sentence are not yet connected by any explicit edge, we add an implicit ``sequence'' edge from the earlier verb to the later one.

\paragraph{AMR-based time roles.}
We further inspect AMR \texttt{:time} roles. If an action node has a \texttt{:time} edge to a time node whose concept is ``after'' or ``before'', we follow the \texttt{:op1} edge of this time node to find a referenced action and connect the two accordingly. For time nodes with concept ``then'', we approximate the ordering from AMR coordination: if the current action is attached to an \emph{and}-node as \texttt{:op2}, and there is another action attached as \texttt{:op1}, we treat the latter as the predecessor of the former.

All temporal edges are stored as quadruples (source, target, type, connective). We then deduplicate them and resolve conflicts: explicit dependency-based edges have highest priority, followed by AMR time edges, and finally purely implicit edges. At most one directed edge is kept for each unordered action pair.

\subsubsection{Action Sequencing}

Given the set of actions and temporal edges, we construct a directed graph where nodes are actions and edges point from earlier to later actions. For edges whose connectives imply an inverted textual order (e.g., ``after'', ``since'', ``once'', ``following'', ``upon''), we flip the direction during graph construction so that the edge always encodes temporal precedence.

We then apply Kahn’s algorithm to obtain a topological order. At each iteration, we select nodes with in-degree zero (sorted by ID for determinism), append them to the output list, and decrement the in-degree of their successors. The order in which actions are removed from the graph defines their temporal indices, which we store as \texttt{temporal\_order}. For convenience, we also attach a lightweight \texttt{temporal\_relation} field to each action, describing the type and connective of one outgoing edge to its next action, when available.

If not all nodes can be removed (indicating cycles or mutually inconsistent constraints), we assign \texttt{temporal\_order} $=-1$ to the remaining actions and append them at the end of the list. The final output is the sorted action sequence $A$, which provides both fine-grained motion attributes and an explicit temporal skeleton that we use in downstream evaluation.

\begin{figure*}[t]
\centering
\includegraphics[width=0.8\linewidth]{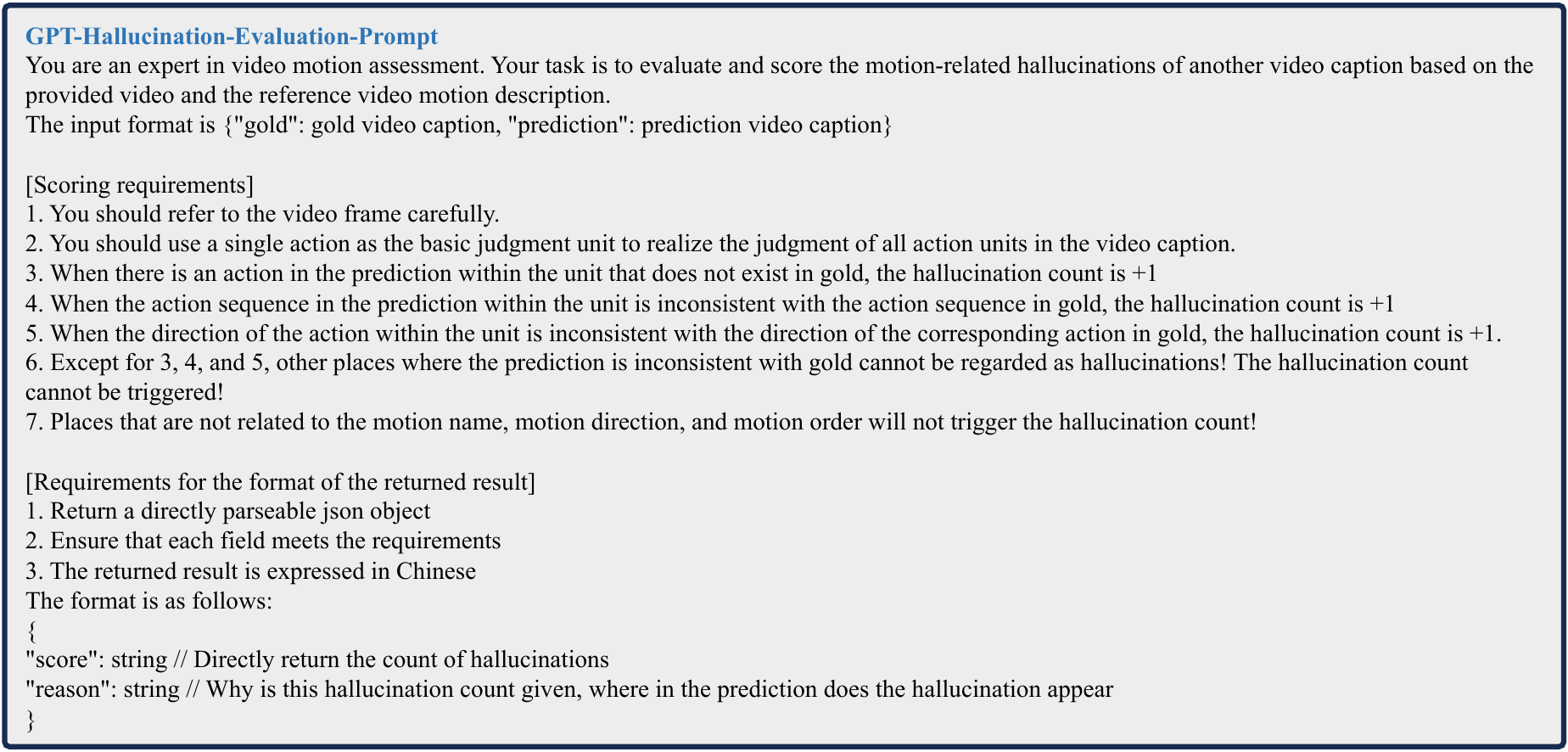}
\caption{
Prompt of hallucination evaluation via GPT-4.1.
}
\label{fig:prompt:hall-gpt-evaluation}
\end{figure*}

\begin{figure*}[t]
\centering
\includegraphics[width=1.0\linewidth]{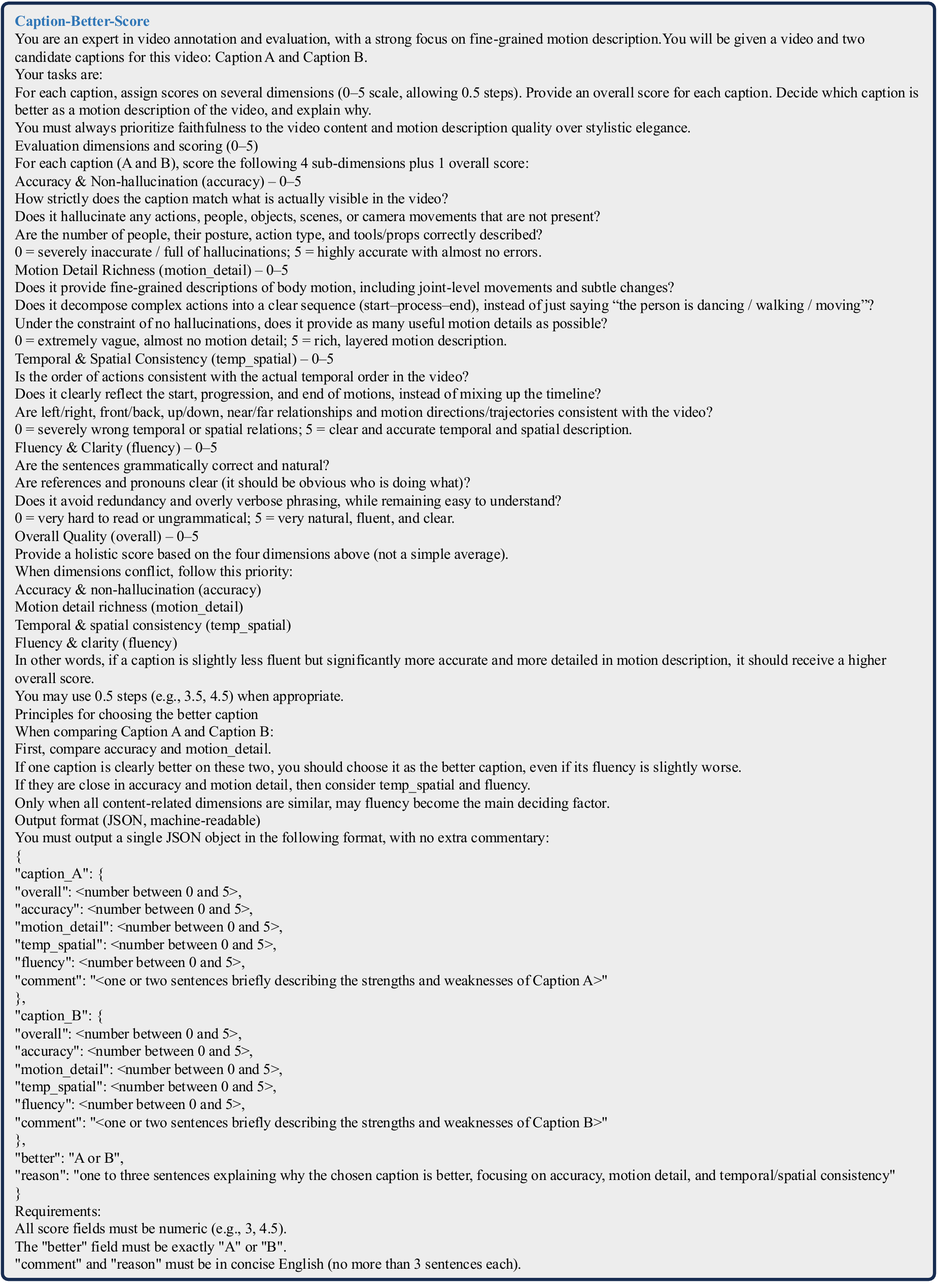}
\caption{
Prompt of Caption Comparison.
}
\label{fig:prompt:gpt-cap-better}
\end{figure*}

\section{QA Design}
In this section, we present the detailed design and construction process of the QA pairs in \benchqa.
First, we apply the pose estimation and kinematic calculation for the videos to get the dense and fine-grained motion-centric video captions in the \benchcap.
Then, we categorize the QA questions into five major attribute types: human attributes, motion attributes, emotional attributes, environmental attributes, and camera-related attributes. 
For each of these categories, we design a set of targeted questions aimed at probing specific sub-attributes, thereby enhancing and evaluating the fine-grained understanding capability of models across these distinct dimensions.

\subsection{Human-related Understanding}
This category assesses the model's ability to perceive and interpret human presence and interaction:
\begin{itemize}
    \item Number of people visible in the video.
    \item Number of people actively engaged in movement.
    \item Whether any interpersonal interactions occur; if so, a description is required.
    \item Whether all individuals are performing the same movement or displaying variations.
\end{itemize}

\subsection{Movement-related Analysis}
This category focuses on both general and fine-grained understanding of movements.

\textbf{(a) Movement Attributes:}
\begin{itemize}
    \item Whether the movement is performed individually or collaboratively.
    \item The type of movement performed (e.g., walking, running, etc.).
    \item Duration of each movement.
    \item Speed or pace of the movement (e.g., fast, moderate, slow).
    \item Whether movements are continuous or intermittent.
    \item Presence of rhythmic or patterned movements.
    \item Consistency in balance or posture during movement.
\end{itemize}

\textbf{(b) Movement Descriptions:}
\begin{itemize}
    \item Detailed description of movements performed by the individual(s).
    \item Observation of subtle movements or gestures.
    \item Body-part specific queries, including movements of legs, hands, and head.
    \item Whether movements are synchronized across different body parts.
    \item Identification of the initiating body part in a movement sequence.
\end{itemize}

\textbf{(c) Temporal Order of Movements:}
\begin{itemize}
    \item Determination of the chronological order in which movements occur.
\end{itemize}

\textbf{(d) Interaction with Objects and People:}
\begin{itemize}
    \item Identification of objects held in hands, including left and right hand specifics.
    \item Analysis of physical interactions with objects or other people (e.g., touching, grabbing, pushing).
    \item Description of object manipulation and its manner.
\end{itemize}

\textbf{(e) Repetition of Movements:}
\begin{itemize}
    \item Counting the number of repetitions of a specific movement within the video.
\end{itemize}

\subsection{Camera Motion Understanding}
This category investigates how camera dynamics affect motion interpretation:
\begin{itemize}
    \item Description of camera movement within the scene.
    \item Characterization of camera motion over time.
    \item Determination of whether the camera is static or dynamic.
    \item Specific attention to camera behavior during key actions (e.g., zooming, focus adjustment).
\end{itemize}

\subsection{Environmental Context}
These questions examine the spatial context in which movement occurs:
\begin{itemize}
    \item Determination of whether the scene is set indoors or outdoors.
    \item Analysis of how the surrounding environment influences the movement.
\end{itemize}

\subsection{Emotional and Expressive Cues}
This section explores the affective dimension of video understanding:
\begin{itemize}
    \item Identification of emotions conveyed through body movements or expressions.
    \item Observation of facial expressions that accompany the movements.
\end{itemize}

This question set supports both factual and descriptive answering and is designed to facilitate fine-grained analysis of visual-language models in human-centric video understanding tasks.

\section{Experiment Details}
\subsection{Training Details}
In this section, we provide additional training details that could not be included in the main paper due to space constraints.
During the pose estimation stage, each video clip is uniformly sampled into 32 frames and annotated using RTMPose3D.

For the SFT stage, we set the frame rate to FPS = 4 and FPS\_MAX\_FRAMES = 64, and train for 1 epoch.
In the GRPO stage, we retain the same FPS and frame limit settings, and train for 2 epochs.
\begin{table*}[t]
\centering
\scalebox{0.8}{
\begin{tabular}{cc|ccccccc}
\toprule
\textbf{Dataset} & \textbf{Version} & \textbf{Better-GPT$\uparrow$} & \textbf{Overall$\uparrow$} & \textbf{Accuracy$\uparrow$} & \textbf{Motion Detail$\uparrow$} & \textbf{Temp Spatial$\uparrow$} & \textbf{Fluency$\uparrow$} & \textbf{Better-Human$\uparrow$}\\
\hline
\multirow{2}{*}{MotionBench} & Our & $47.7\%$ & $4.095$ & $3.945$ & $4.313$ & $3.891$ & $4.158$ & $49.2\%$\\
 & Original & $52.3\%$ & $4.421$ & $4.012$ & $4.132$ & $3.924$ & $4.203$ & $50.8\%$\\
\hline
\multirow{2}{*}{MoVid} & Our & $71.6\%$ & $4.285$ & $4.462$ & $4.278$ & $4.351$ & $4.372$ & $57.9\%$\\
 & Original & $28.4\%$ & $3.692$ & $3.984$ & $3.675$ & $3.928$ & $4.195$ & $42.1\%$\\
\hline
\multirow{2}{*}{Dream-1K} & Our & $69.8\%$ & $4.267$ & $4.438$ & $4.261$ & $4.337$ & $4.358$ & $55.7\%$\\
 & Original & $30.2\%$ & $3.714$ & $3.995$ & $3.492$ & $3.945$ & $4.211$ & $44.3\%$\\
\bottomrule
\end{tabular}
}
\caption{The comparison between KPM version caption and other dataset caption.
The numerical value ranges from 1 to 5.}
\label{tab:supp:exp:cap:compare}
\end{table*}
\subsection{Baselines}
\textbf{Closed-source VLM APIs.} 
GPT-4.1 is one of the most powerful general-purpose closed-source APIs currently available.
Gemini-2.5-Pro is another closed-source VLM with powerful video understanding capabilities.
To ensure a fair comparison with our model, we carefully designed a high-quality prompt for GPT and Gemini as shown in Section 2.2 of Supplementary Material, guiding them to generate high-quality motion-centric fine-grained captions, rather than producing very brief and ambiguous prompt.

\textbf{Open-source VLMs.} 
Qwen2-VL introduce the naive dynamic resolution and Multimodal Rotary Position Embedding (M-RoPE) techniques to process images and videos of varying resolutions and lengths.
Qwen2.5-VL incorporates dynamic resolution processing and absolute time encoding based on Qwen2-VL.
InternVideo2.5 improve the ability to understand long videos and fine-grained visual perception are significantly via introducing long and rich context modeling and hierarchical token compression technology.
MiniCPM-V optimizes parameter efficiency for lightweight deployment while maintaining competitive video QA performance.
VideoLLaMA3 achieve efficient understanding of images and videos through high-quality image-text data training and multi-stage fine-tuning strategies.
Tarsier2-Recap is the re-caption version of Tarsier2, a VLM which utilizes Direct Preference Optimization (DPO) techniques to excel in tasks such as video description, question answering, and localization.

\section{Other Experiments}
\begin{figure*}[t]
\centering
\includegraphics[width=0.95\linewidth]{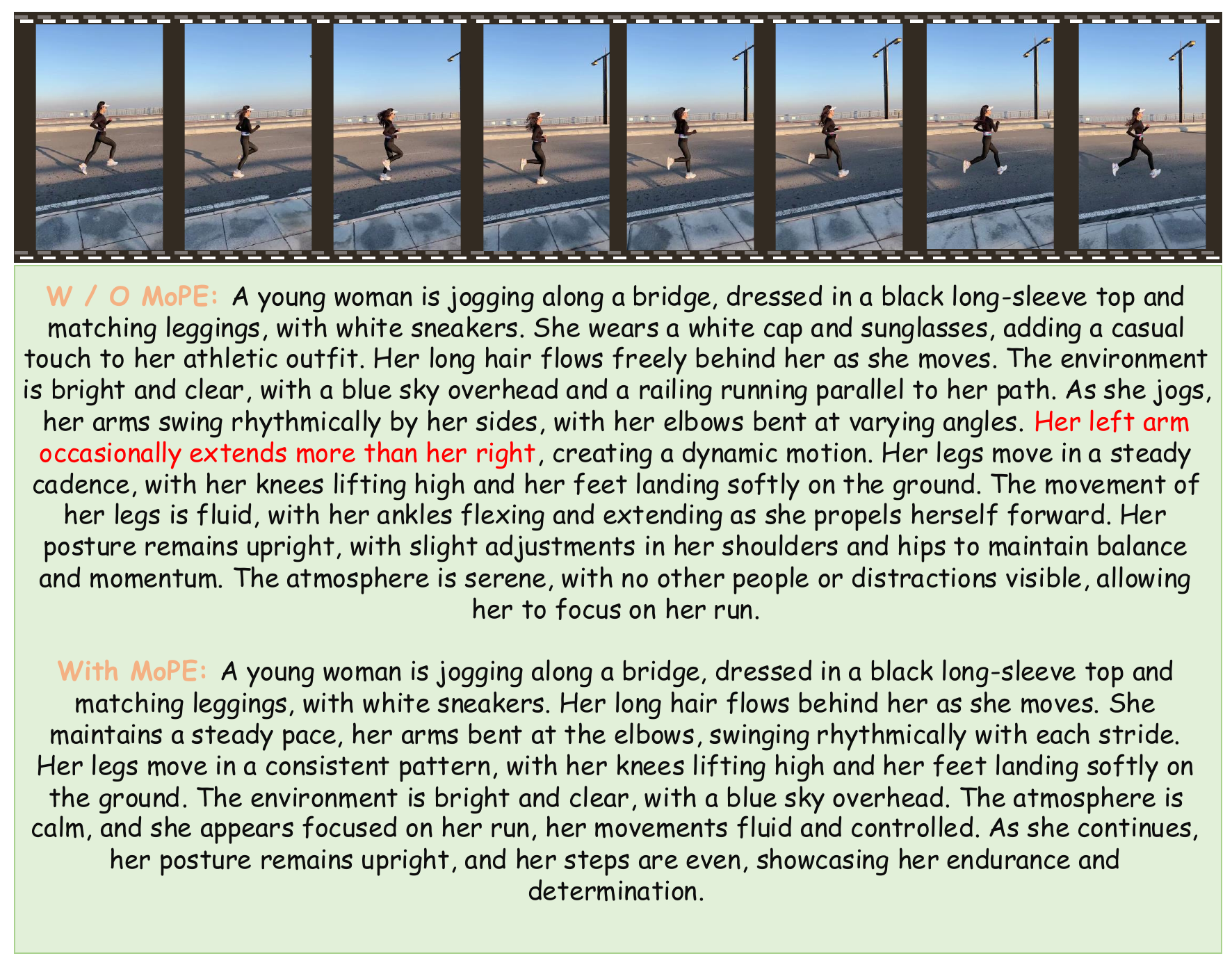}
\caption{
The visualization case of hallucination between KPM and KPM w / o MoPE.
The red text corresponds to the hallucination content.
}
\label{fig:case:hall}
\end{figure*}

\begin{table}[t]
\centering
\scalebox{0.6}{
\begin{tabular}{c|cccccc}
\toprule
\textbf{Version} & \textbf{Better$\uparrow$} & \textbf{Overall$\uparrow$} & \textbf{Accuracy$\uparrow$} & \textbf{Motion Detail$\uparrow$} & \textbf{Temp Spatial$\uparrow$} & \textbf{Fluency$\uparrow$}\\
\hline
KPM-3D & $80.25\%$ & $4.309$ & $4.519$ & $4.319$ & $4.393$ & $4.394$ \\
KPM-2D & $19.75\%$ & $3.736$ & $4.019$ & $3.699$ & $3.962$ & $4.226$ \\
\bottomrule
\end{tabular}
}
\caption{The comparison between 3D version and 2D version of KPM.
The numerical value ranges from 1 to 5.}
\label{tab:supp:exp:2dv3d}
\end{table}
\begin{table}[t]
\centering
\scalebox{0.6}{
\begin{tabular}{ccccccc}
\toprule
\textbf{Vel$\uparrow$} & \textbf{Angle-Vel$\uparrow$} & \textbf{Vel-Score$\uparrow$} & \textbf{Angle-Score$\uparrow$} & \textbf{Frame-Angles$\uparrow$} & \textbf{Vel-FFT$\uparrow$} & \textbf{Angle-FFT$\uparrow$}\\
\hline
$3.637$ & $3.821$ & $4.577$ & $3.766$ & $3.485$ & $3.883$ & $3.975$ \\
\bottomrule
\end{tabular}
}
\caption{Ablation study of kinematic calculation. 
The numerical value ranges from 1 to 5.}
\label{tab:supp:exp:ablation:kinematic}
\end{table}

\subsection{Ablation Study on Pose Estimation Dimension}
As a comparison, in addition to using 3D pose estimation, we also employ AlphaPose to achieve 2D pose estimation and construct 2D version of KPM captions based on these results.
Specifically, we feed the 2D keypoints into the same kinematic calculation pipeline and apply the same captioning procedure to obtain 2D-pose-based motion captions. 
We then use GPT-4.1 as an automatic judge, guided by the prompt shown in Fig~\ref{fig:prompt:gpt-cap-better}, to compare the quality of the 3D-based KPM captions against their 2D-based counterparts. 
The evaluation results are shown in Table~\ref{tab:supp:exp:2dv3d}.

\subsection{Ablation Study on Kinematic Calculation}
As described in Section 3.1 of the main paper, our kinematic calculation pipeline yields a variety of kinematic-related attributes that characterize the underlying motion process. 
To examine how each attribute contributes to the quality of the KPM-Caption, we design a set of ablation experiments that selectively remove individual attributes. The results are presented in Table~\ref{tab:supp:exp:ablation:kinematic}.

\subsection{Cross-Dataset Evaluation of KPM Captions}
As described in Section 3.5 of the main paper, we further apply our KPM pipeline to re-annotate several existing motion-centric video datasets, generating KPM-style motion captions for these videos. 
We then evaluate the resulting captions using the same protocol as in Section 5.1 of the Supplementary Material, and report the quantitative results in Table~\ref{tab:supp:exp:cap:compare}.
In addition, we also conduct a user study to the resulting captions. 
The user study results are also reported in Table~\ref{tab:supp:exp:cap:compare}.

Consistent with sections 5.1 and 5.2 of the Supplementary Material, the score range is also 1-5. 
It can be seen that the original caption from MotionBench is slightly better than KPM-Caption in overall quality. 
This is because MotionBench employs rigorous multi-round manual annotation and also provides detailed descriptions of fine-grained motion. 
While Dream-1K also uses manual annotation, as shown in Section 3.5 of the main paper, its overly brief description causes the caption to lose a significant number of points in terms of motion detail.

\subsection{Limitation and Future Work}
A key limitation of this work is that our framework is explicitly tailored to human motion, rather than general motion phenomena. 
This design choice is reflected in our dataset construction, annotation protocol, and evaluation metrics, all of which assume articulated human bodies as the primary subjects. 
As a result, the proposed method and benchmark do not achieve better performance on other types of motion, such as animal behaviors, vehicle dynamics, or non-rigid object deformations, and their applicability to these broader domains remains an open question.
To better contextualize the impact of our contributions, we also highlight several concrete application scenarios where fine-grained human motion understanding is particularly valuable. 
For instance, our framework could support sports analytics by providing detailed, interpretable descriptions of athletes’ movements; it could facilitate physical therapy and rehabilitation monitoring by characterizing deviations from desired motion patterns over time; and it could enhance human–robot interaction by enabling robots to perceive and reason about subtle human motion cues in a more structured way.
We leave the adaptation and validation of our approach in these downstream applications, as well as its extension beyond human motion, as promising directions for future work.

\subsection{Visualization Cases}
\begin{figure*}[t]
\centering
\includegraphics[width=0.95\linewidth]{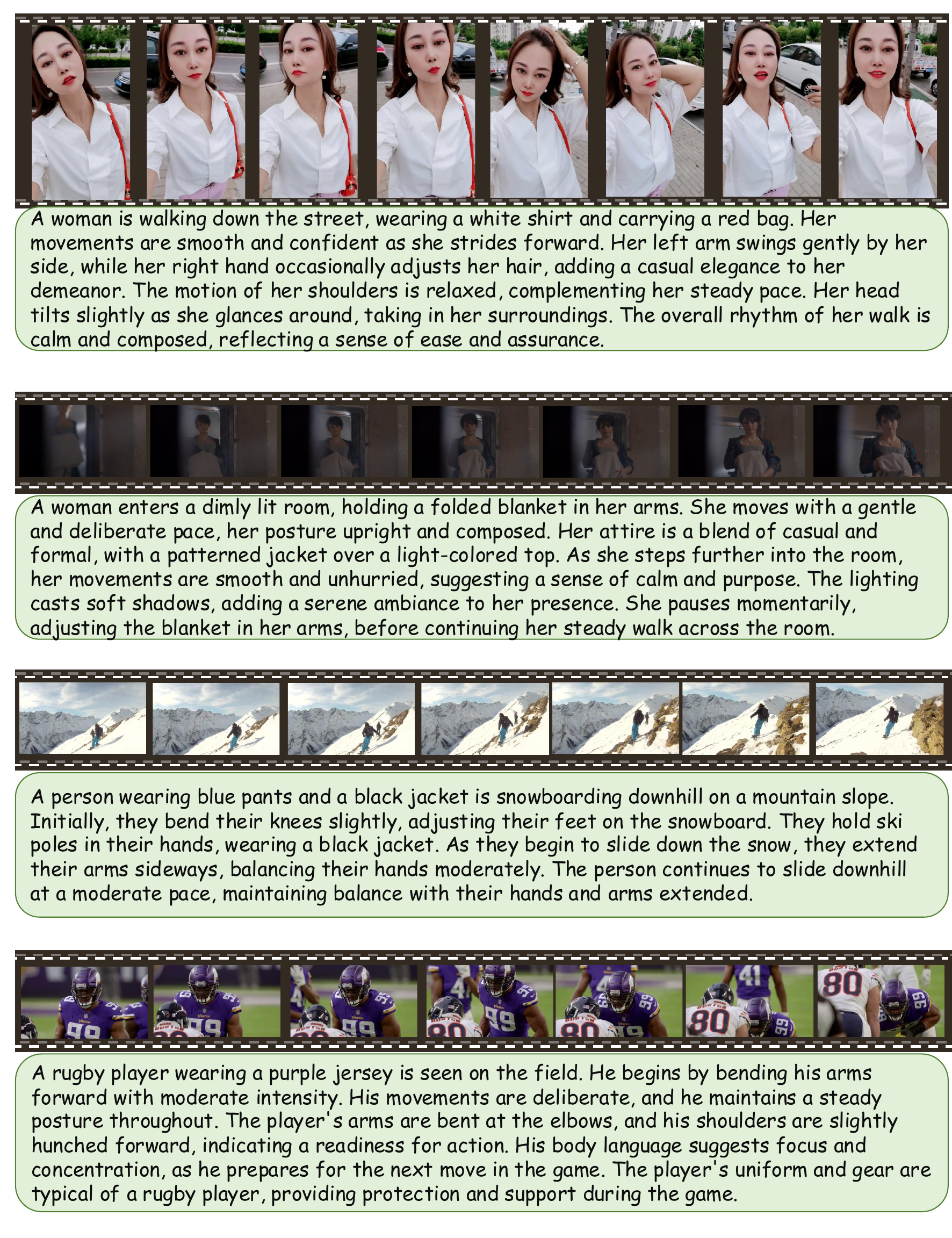}
\caption{
The visualization cases for dense and fine-grained motion-centric video caption.
}
\label{fig:case:caption}
\end{figure*}

\begin{figure*}[t]
\centering
\includegraphics[width=1.0\linewidth]{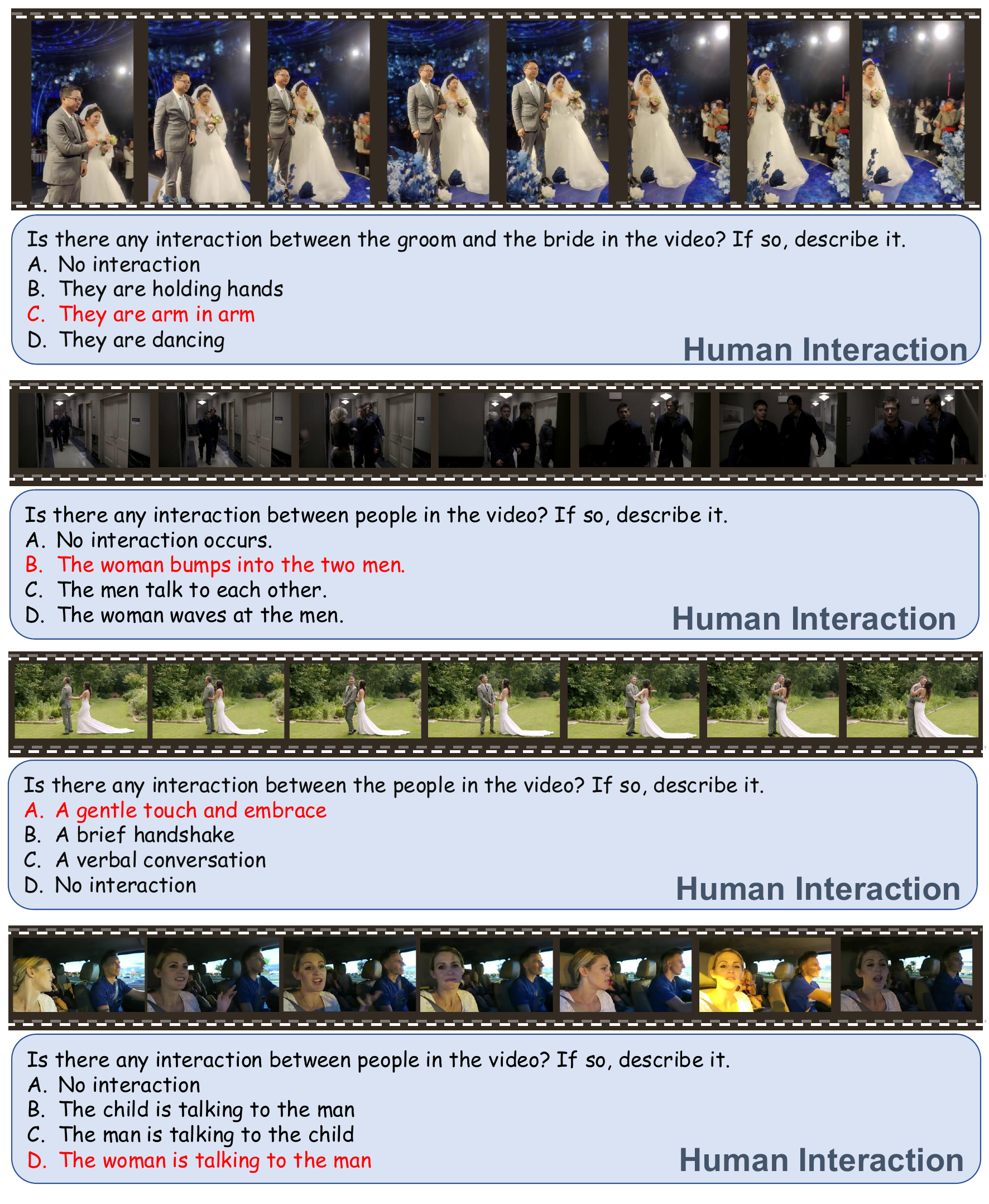}
\caption{
The visualization cases for human interaction type question.
}
\label{fig:case:qa:human:interaction}
\end{figure*}

\begin{figure*}[t]
\centering
\includegraphics[width=1.0\linewidth]{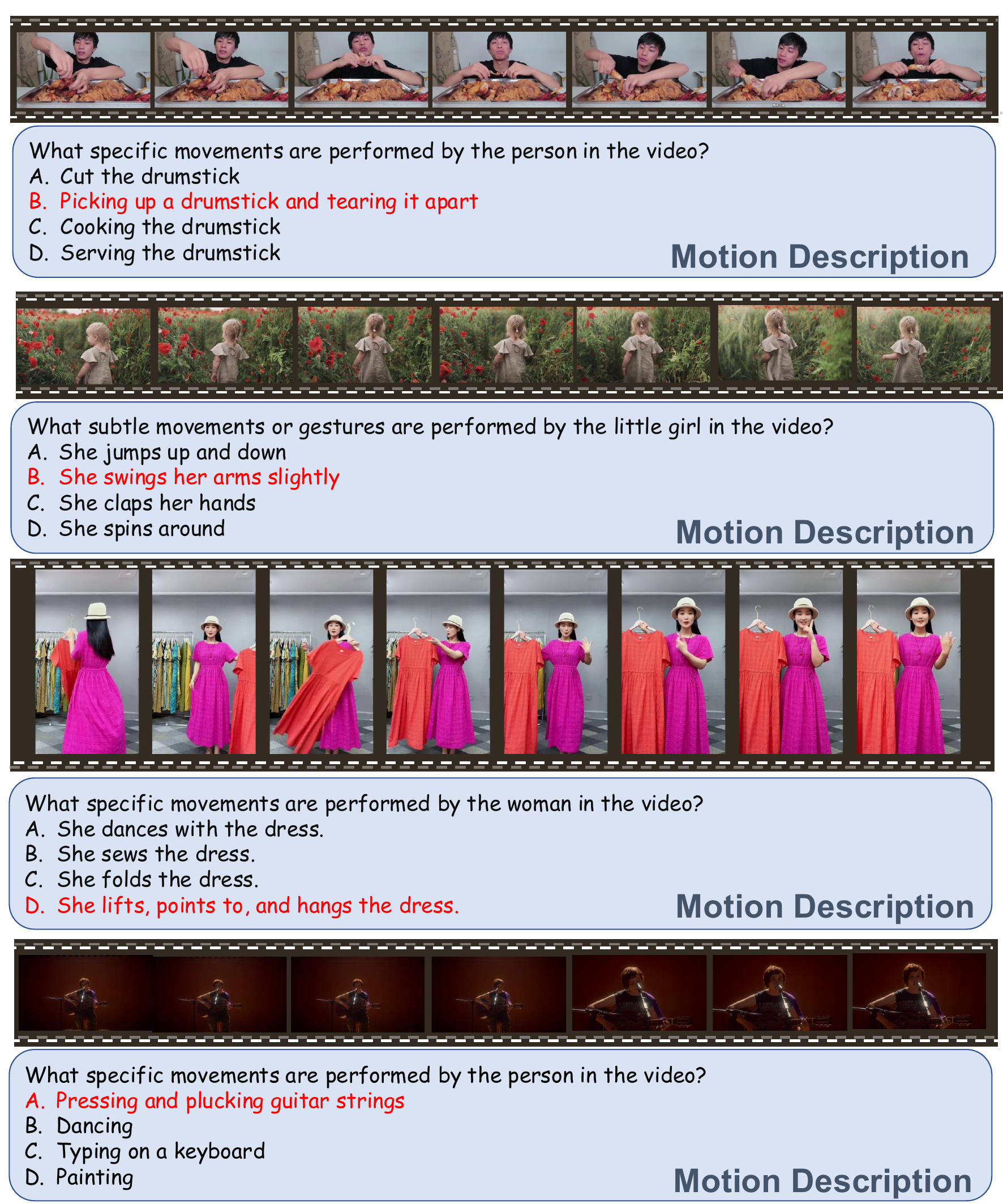}
\caption{
The visualization cases for motion description type question.
}
\label{fig:case:qa:move:desc}
\end{figure*}

\begin{figure*}[t]
\centering
\includegraphics[width=1.0\linewidth]{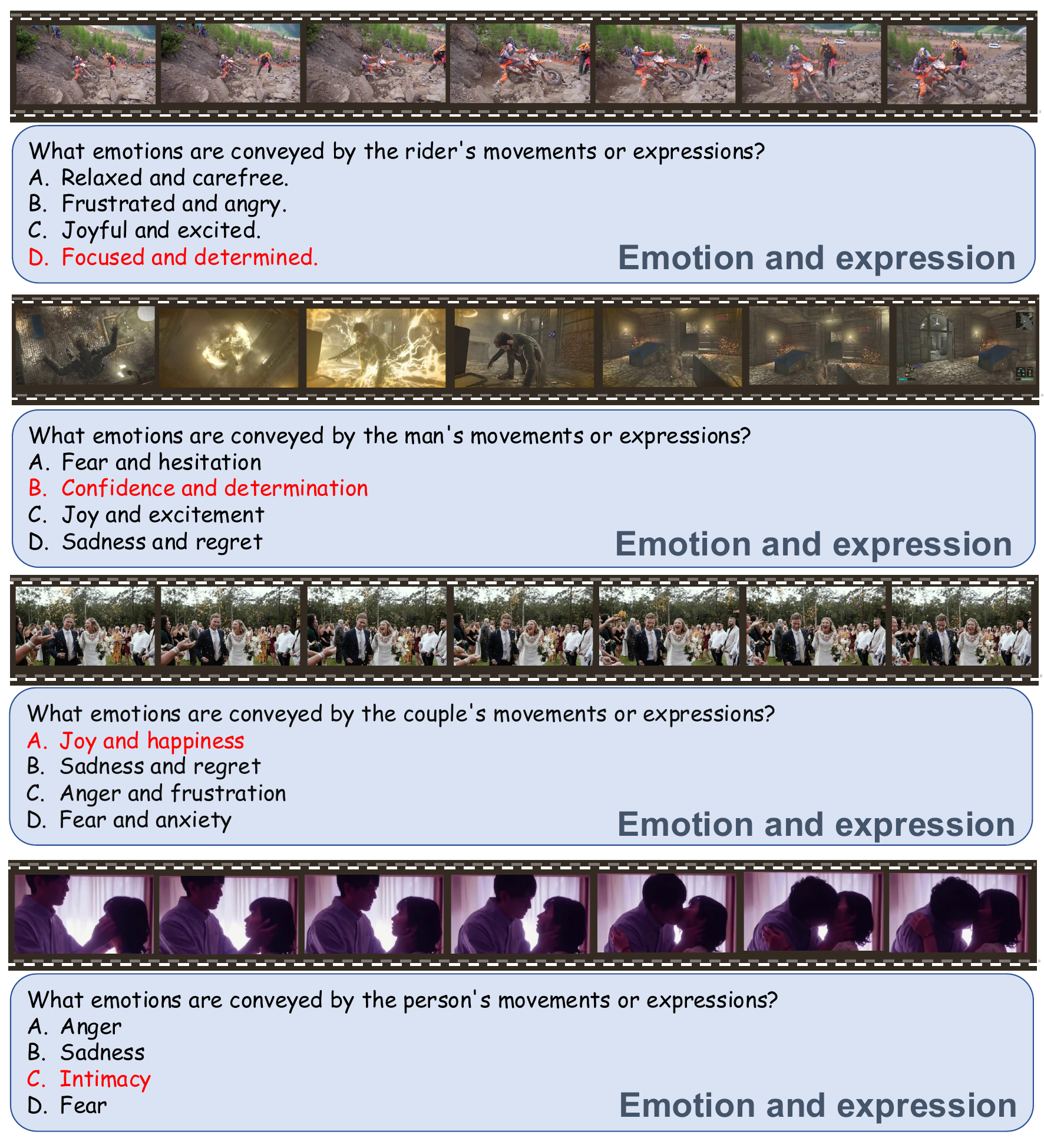}
\caption{
The visualization cases for motion emotion type question.
}
\label{fig:case:qa:emo}
\end{figure*}

\begin{figure*}[t]
\centering
\includegraphics[width=1.0\linewidth]{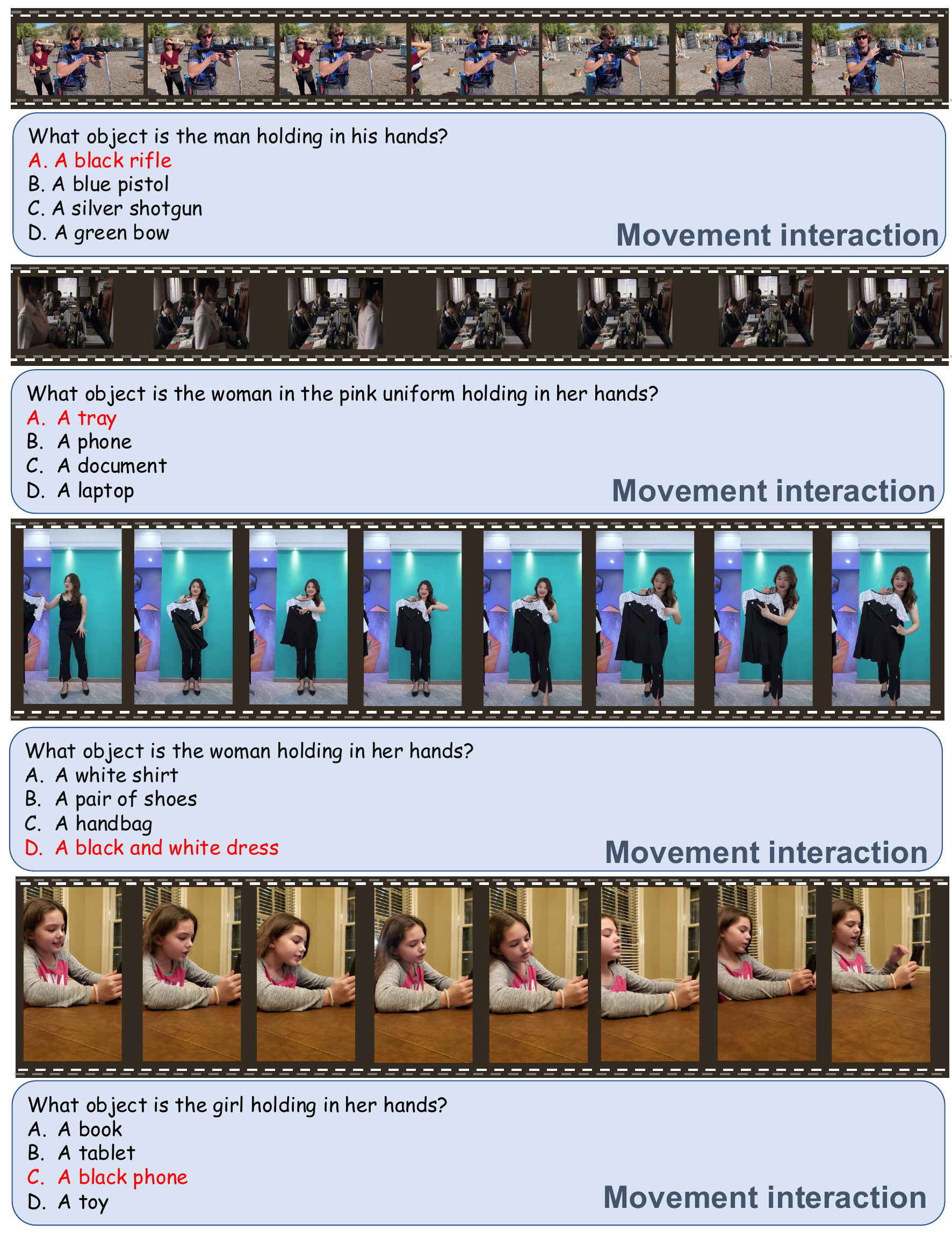}
\caption{
The visualization cases for movement interaction type question.
}
\label{fig:case:qa:move:interaction}
\end{figure*}

This section presents several visualization cases of our work.
The Figure~\ref{fig:case:hall} shows a comparison of caption hallucination before and after KPM training with GRPO within MoPE. The red text indicates the motion-related hallucination.
The Figure~\ref{fig:case:caption} presents the dense and fine-grained motion-centric video caption.
The Figure~\ref{fig:case:qa:human:interaction} presents case examples for human interaction type question.
The Figure~\ref{fig:case:qa:move:desc} presents case examples for motion description type question.
The Figure~\ref{fig:case:qa:emo} presents case examples for motion emotion type question.
The Figure~\ref{fig:case:qa:move:interaction} presents case examples for movement interaction type question.

\end{document}